\documentclass[sigconf]{acmart}
\usepackage{amsmath,amsfonts}
\usepackage{graphicx}
\usepackage{booktabs}
\usepackage{textcomp}
\usepackage{xcolor}
\usepackage{float}
\usepackage{caption}
\usepackage{subfig}
\usepackage{bm}
\usepackage{multirow}
\usepackage{amsthm}
\usepackage{mathrsfs}
\usepackage[noend]{algpseudocode}
\usepackage{graphicx}  

\usepackage[ruled,vlined]{algorithm2e}
\AtBeginDocument{%
  \providecommand\BibTeX{{%
    \normalfont B\kern-0.5em{\scshape i\kern-0.25em b}\kern-0.8em\TeX}}}

\begin{document}

\title{Make Heterophily Graphs Better Fit GNN: A Graph Rewiring Approach}

\author{Wendong Bi}
\authornote{Work done while this author was an intern at Microsoft Research}
\email{biwendong20@mails.ucas.ac.cn}
\affiliation{%
  \institution{Institute of Computing Technology, Chinese Academy of Sciences}
  \city{Beijing}
  \country{China}
}
\author{Lun Du}
\authornote{Corresponding Author}
\email{lun.du@microsoft.com}
\affiliation{%
  \institution{Microsoft Research Asia}
  \city{Beijing}
  \country{China}}

\author{Qiang Fu}
\email{qifu@microsoft.com}
\affiliation{%
  \institution{Microsoft Research Asia}
  \city{Beijing}
  \country{China}}

\author{Yanlin Wang}
\email{yanlwang@microsoft.com}
\affiliation{
  \institution{Microsoft Research Asia}
  \city{Beijing}
  \country{China}}

\author{Shi Han}
\email{shihan@microsoft.com}
\affiliation{
  \institution{Microsoft Research Asia}
  \city{Beijing}
  \country{China}}
  
\author{Dongmei Zhang}
\email{dongmeiz@microsoft.com}
\affiliation{%
  \institution{Microsoft Research Asia}
  \city{Beijing}
  \country{China}}

\begin{abstract}
		Graph Neural Networks (GNNs) are popular machine learning methods for modeling graph data. A lot of GNNs perform well on homophily graphs while having unsatisfied performance on heterophily graphs. Recently, some researchers turn their attentions to designing GNNs for heterophily graphs by adjusting message passing mechanism or enlarging the receptive field of the message passing. Different from existing works that mitigate the issues of heterophily from model design perspective, we propose to study heterophily graphs from an orthogonal perspective by rewiring the graph structure to reduce heterophily and making the traditional GNNs perform better. Through comprehensive empirical studies and analysis, we verify the potential of the rewiring methods. To fully exploit its potential, we propose a method named \textbf{D}eep \textbf{H}eterophily \textbf{G}raph \textbf{R}ewiring (DHGR) to rewire graphs by adding homophilic edges and pruning heterophilic edges. The detailed way of rewiring is determined by comparing the similarity of \textbf{label/feature-distribution of node neighbors}. Besides, we design a scalable implementation for DHGR to guarantee a high efficiency. DHRG can be easily used as a plug-in module, i.e., a graph pre-processing step, for any GNNs, including both GNN for homophily and heterophily, to boost their performance on the node classification task. To the best of our knowledge, it is the first work studying graph rewiring for heterophily graphs. Extensive experiments on 11 public graph datasets demonstrate the superiority of our proposed methods.
	\end{abstract}


\maketitle
	\section{Introduction}
	Graph-structure data is ubiquitous in representing complex interactions between objects \cite{du2018traffic,chen2020TSSRGCN,song2020inferring}. Graph Neural Network (GNN), as a powerful tool for graph data modeling, has been widely developed for various real-world applications \cite{du2021tabularnet,yao2022trajgat,wang2020cocogum}. Based on the message passing mechanism, GNNs update node representations by aggregating messages from neighbors, thereby concurrently exploiting the rich information inherent in the graph structure and node attributes. 
	
	Traditional GNNs \cite{GCN,GAT,GraphSAGE} mainly focus on \textbf{homophily} graphs that satisfy property of homophily (i.e. most of connected nodes belong to the same class).
	However, these GNNs usually can not perform well on graphs with \textbf{heterophily} (i.e. most of connected nodes belong to different classes) for the node classification problem, because message passing between nodes from different classes makes their representations less distinguishable, and thus leading to bad performance on node classification task.
	The aforementioned issues motivate considerable studies around GNNs for heterophily graph. 
	For example, some studies \cite{wang2021powerful,du2022gbk,yan2021two} adjust message passing mechanism for heterophily edges,  while others \cite{abu2019mixhop,h2gcn,chien2020adaptive,pei2020geom} enlarge the receptive field for the message passing. Note that, all these works mitigate the distinguishability issue caused by heterophily from the perspective of the GNN model design. While, there is another orthogonal perspective to mitigate the issue caused by heterophily, i.e., rewiring graph to reduce heterophily or increase homophily, which is still under-explored. 

	Graph rewiring~\cite{alon2020bottleneck, topping2021understanding, LDS, chen2020iterative} is a kind of method that decouples the input graph from the graph for message passing and boost the performance of GNN on node classification tasks via changing the message passing structure. Many works have utilized graph rewiring for different tasks. However, most existing graph rewiring techniques have been developed for graphs under homophily assumption (sparsity~\cite{sparse}, smoothness~\cite{smooth1, smooth2} and low-rank~\cite{lowrank}), and thereby can not directly transfer to heterophily graphs. Different from existing solutions that design specific GNN architectures adapted to heterophily graphs, in this paper, we conduct comprehensive study on graph rewiring and propose an effective rewiring algorithm to reduce graph heterophily, which make GNNs perform better for both heterophily and homophily graphs.

	First we demonstrate the effects of increasing homophily-level  for heterophily graphs in Sec.~\ref{section:findings} with comprehensive controlled experiments. Note that the homophily (and heterophily) level can be measured with Homophily Ratio (HR) \cite{pei2020geom,h2gcn}, which is formally defined as an average of the consistency of labels between each connected node-pair. From the analysis in Sec.~\ref{subsec:finding_hr_deg}, we find that both the \textbf{node-level homophily ratio} \cite{du2022gbk,pei2020geom} and \textbf{node degree} (reflects the recall of nodes from the same class) can affect the performance of GCN on the node classification task, where increasing either of the two variables can lead to better performance of GCN. This finding, i.e., classification performance of GCN on heterophily graphs can be increased by reducing the heterophily-level of graphs, motivates us to design a graph-rewiring strategy to increase homophily-level for heterophily graphs so that GNNs can perform better on the rewired graphs.

	Then, we propose a learning-based graph rewiring approach on heterophily graphs, namely \textbf{D}eep \textbf{H}eterophily \textbf{G}raph \textbf{R}ewiring (DHGR). DHGR rewires the graph by adding/pruning edges on the input graph to reduce its heterophily-level. It can be viewed as a plug-in module for graph pre-processing that can work together with many kinds of GNN models including both GNN for homophily and heterophily, to boost their performance on node classification tasks. 
	\begin{figure}[h]
		\centering
		\includegraphics[width=\linewidth]{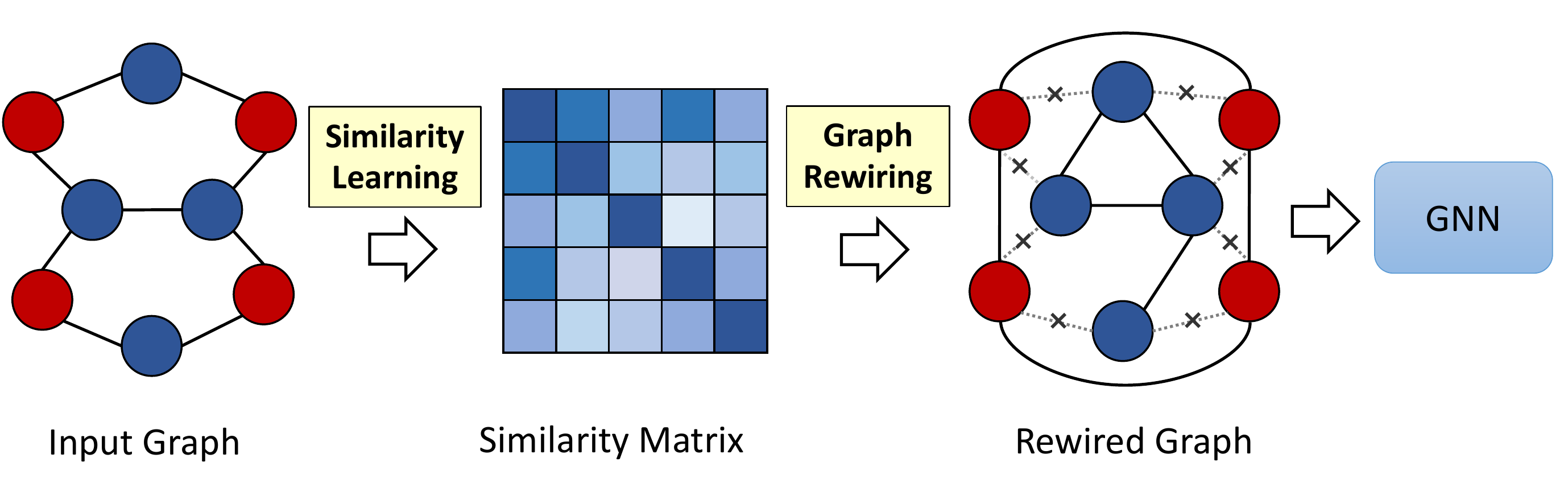}
		\caption{Pipeline of Graph Rewiring for heterophily graphs. Red and blue circles denote nodes from different classes. }
		\label{fig:piepline}
	\end{figure}
	The key idea of DHGR is to reduce the heterophily while keeping the effectiveness by adding more homophilic edges and removing heterophilic edges. 
	However, simply adding homophilic edges and removing heterophilic edges between nodes in the training set may increase the risk of overfitting and lead to poor performance (we prove this in Sec.~\ref{sec:main_res}). 
	Another challenge is that unlike homophily graphs that can leverage Laplace Smooth to enhance the correlation between node features and labels, heterophily graphs do not satisfy the property of smoothness~\cite{smooth2, smooth1}.
	In this paper, we propose to use \textbf{label/feature-distribution of neighbors} on the input graph as guidance signals to identify edge polarity (homophily/heterophily)  and prove its effectiveness in ~\ref{subsec:finding_mi}.
	
	Under the guidance of the neighbors' label-distribution, DHGR learns the similarity between each node-pair, which forms a similarity matrix. Then based on the learned similarity matrix, we can rewire the graph by adding edges between high-similarity node-pairs and pruning edges connecting low-similarity node-pairs. Then the learned graph structure can be further fed into GNNs for node classification tasks. Besides, we also design a scalable implementation of DHGR which avoids the quadratic time and memory complexity with respect to the numbers of nodes, making our method available for large-scale graphs.  Finally, extensive experiments on 11 real-world graph datasets, including both homophily and heterophily graphs, demonstrate the superiority of our method.
	
	We summarize the contributions of this paper as follows:
	\begin{enumerate}
		\item We propose the new perspective, i.e., graph rewiring, to deal with heterophily graphs by reducing heterophily and make GNNs perform better.
		\item We propose to use neighbor's label-distribution as guidance signals to identify homophily and heterophily edges with comprehensive experiments. 
		\item We design a learnable plug-in module for graph rewiring on heterophily graphs, namely DHGR. And we design a high-efficient scalable training algorithm for DHGR. 
		\item We conduct extensive experiments on 11 real-world graphs, including both heterophily and homophily graphs. The results show that GNNs with DHGR consistently outperform their vanilla versions.  In addition, DHGR has additional gain even when combined with GNNs specifically designed for heterophily graphs.
	\end{enumerate}

	\begin{figure*}[h]
		\begin{minipage}[t]{0.32\linewidth}
			\centering
			\subfloat[Cora.]{\includegraphics[width=\linewidth]{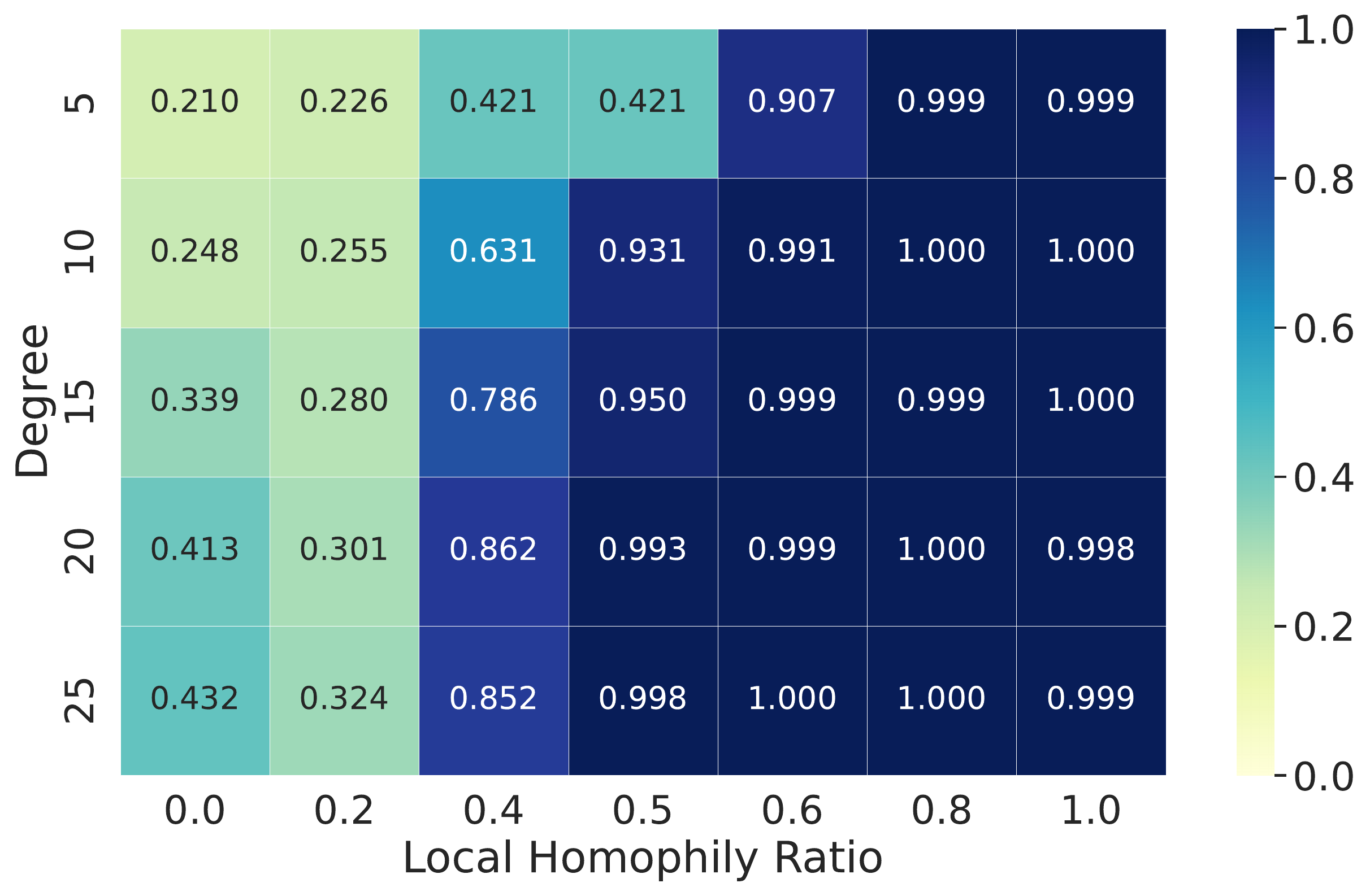}}
		\end{minipage}%
		\begin{minipage}[t]{0.32\linewidth}
			\centering
			\subfloat[Chameleon]{\includegraphics[width=\linewidth]{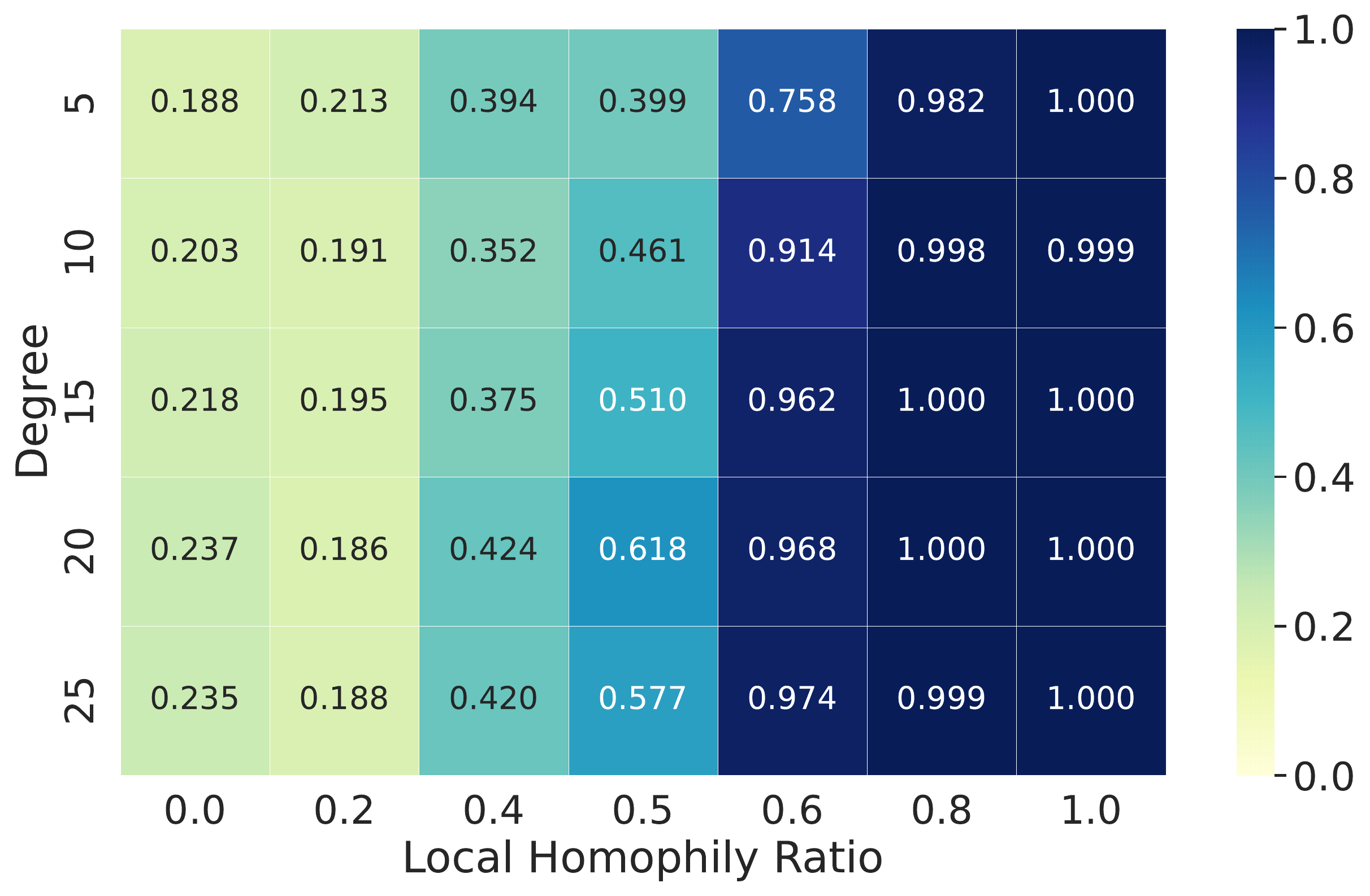}}
		\end{minipage}
		\begin{minipage}[t]{0.32\linewidth}
			\centering
			\subfloat[Actor]{\includegraphics[width=\linewidth]{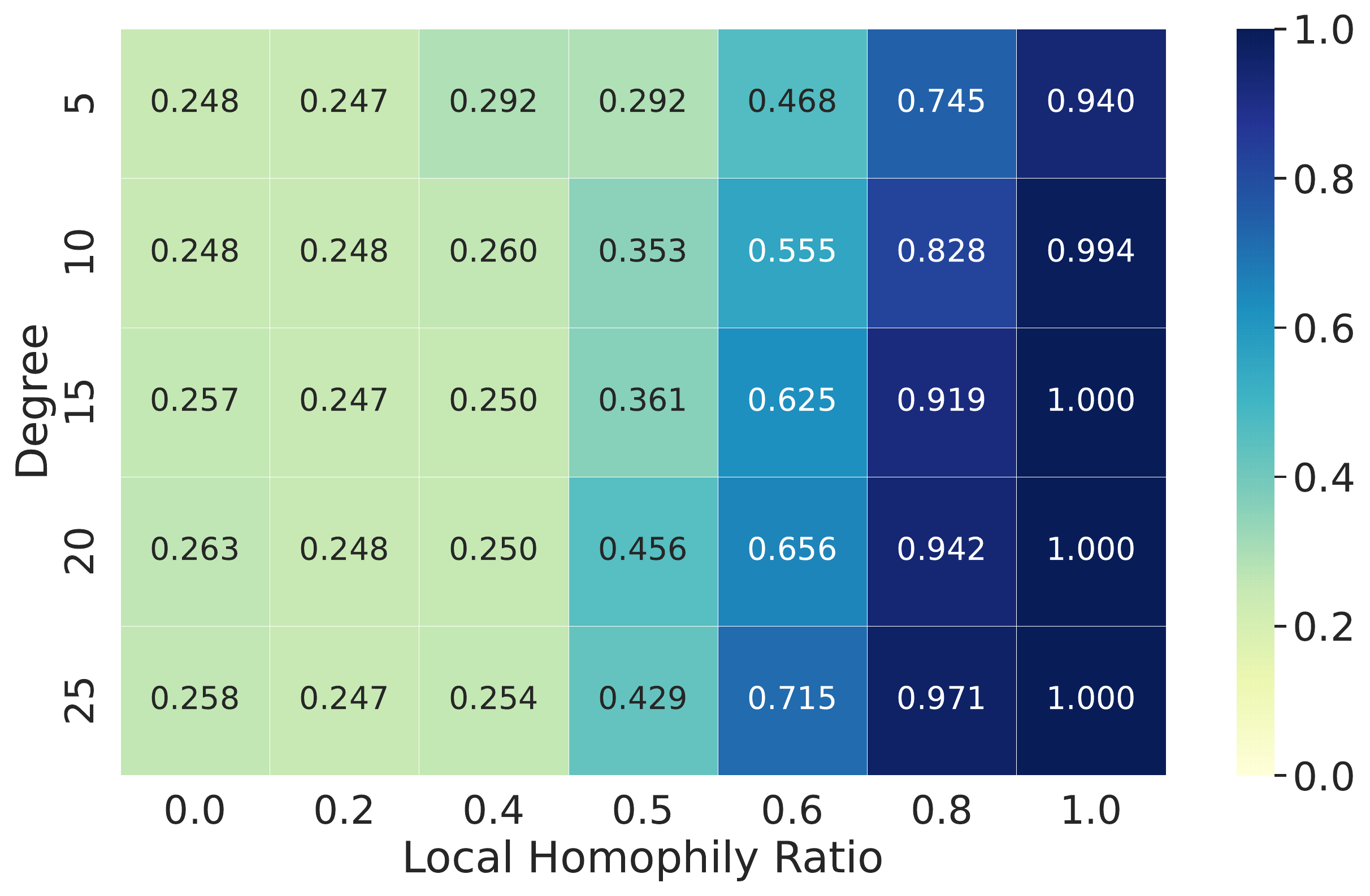}}
		\end{minipage}
		\caption{Graph rewiring validation experiments on three datasets. Each block in the heatmap denotes a rewired graph with $d$ node degree and $p$ node-level homophily ratio. The values in the block denote node classification accuracy of vanilla GCN on the test set (average accuracy of 3 runs).}
		
		\label{fig:gr_valid}
	\end{figure*}
\section{Preliminary}
	In this section, we give the definitions of some important terminologies and concepts appearing in this paper.
	\subsection{Graph Neural Networks}
	Let $\mathcal{G}=(V,E)$ denotes a graph, where $V$ is the node set, $N=|V|$ is the number of nodes in $\mathcal{G}$. Let $X\in \mathbb{R}^{N\times D}$ denote the feature matrix and the $l$-th row of $X$ denoted as $x_i$ is the $D$-dimensional feature vector of node $v_i$. $E=\{(v_i, v_j) | v_i, v_j \in V$ and $v_i, v_j$ is connected$\}$. GNNs aim to learn representation for nodes in the graph. Typically, GNN models follow a neighborhood aggregation framework, where node representations are updated by aggregating information of its neighboring nodes. Let $h_i^{(l)}$ denotes the output vector of node $v_i$ at the $l$-th hidden layer and let $h_i^{(0)}=x_i$. The $l$-th iteration of aggregation step can be written as:
	\begin{equation}
	\nonumber
	h_i^{(l)} = \text{COMBINE}\left(h_i^{(l-1)}, \text{AGG}\big(\{h_j^{(l-1)} | v_j \in \mathcal{N}(v_i) \}\big)\right) 
	\end{equation}
	where $\mathcal{N}(v_i)$  is the set of neighbors of $v_i$. The AGG function indicates the aggregation function aimed to gather information from neighbors and the goal of the COMBINE function is to fuse the information from neighbors and the central node. For graph-level tasks, an additional READOUT function is required to get the global representation of the graph. 
	\subsection{Graph Rewiring}
	Given a graph $\mathcal{G} =(V, E)$ with node features $X\in \mathbb{R}^{N\times D}$ as the input, Graph Rewiring (GR) aims at learning an optimal $\mathcal{G}^* = (V, E^*)$ under a given criterion, where the edge set is updated and the node set is constant. Let $A, A^*\in\mathbb{R}^{N\times N}$ denote the adjacent matrix of $\mathcal{G}$ and $\mathcal{G^*}$, respectively. The rewired graph $\mathcal{G}^*$ is used as input of GNNs, which is expected to be more effective than directly inputting the original graph $\mathcal{G}$. As shown in Fig.~\ref{fig:piepline}, the pipeline of Graph Rewiring models usually involves two stages, the similarity learning and the graph rewiring based on the learned similarity between pairs of nodes. It is obvious that the criterion (i.e., objective function) design plays a critical role for the similarity learning stage. Thus, we first mine knowledge from data in the next section to abstract an effective criterion of graph rewiring.
	\section{Observations from data}
	\label{section:findings}
    We observed from data that there exist two important properties of graph (i.e., \textbf{node-level homophily ratio\footnote{Node-level homophily ratio is the homophily ratio of one specific node, which equals the percent of the same-class neighbors in all neighboring nodes.}} and \textbf{degree}) that are strongly correlated with the performance of GNNs. And the two properties provide vital guidance so that we can optimize the graph structure by graph rewiring. However, we cannot directly calculate node-level homophily ratio because of the partially observable labels during training. Therefore, we introduce two other effective signals, i.e., neighbor's observable label/feature-distribution, which have strong correlations with the node-level homophily ratio. In this section, we first verify the relations between the two properties and the performance of GNNs. Then we verify the correlations between neighbor distribution and node-level homophily ratio.
    \subsection{Effects of Node-level Homophily Ratio and Degree}
	\label{subsec:finding_hr_deg}
	 First, we conduct validation experiments to verify the effects of \textbf{node-level homophily ratio} \cite{pei2020geom, du2021gbk} and \textbf{node degree} on the performance of GCN, as guidance for graph rewiring. Specifically, we first construct graphs by quantitatively controlling the node-level homophily ratio and node degree, and then verify the performance of GCN on the constructed graphs as a basis for measuring the quality of constructed graph structure. Note that considering the direction of message passing is from source nodes to target nodes, the node degree mentioned in this paper refers to the in-degree. For example, given node degree $k$ and node-level homophily ratio $p$, we can constructed a directional Graph $\mathcal{G}_{k,p}$ where each node on the $\mathcal{G}_{k,p}$ has $k$ different neighboring nodes pointing to it and there are $\lfloor p \cdot k\rfloor$ same-class nodes among the k neighbors, with other $\lceil k\cdot (1-p)\rceil$ neighbors randomly selected from remaining different-class nodes on the graph.

	As shown in the Fig.~\ref{fig:gr_valid}, we conduct validation experiments on three different graph datasets, including one homophily graph (Cora) and two heterophily graphs (Chameleon, Actor). In this experiments, we construct graphs $\mathcal{G}_{k, p}$ with node degree $k$ ranging from $5$ to $25$ and node-level homophily ratio $p$ ranging from $0.0$ to $1.0$, totally 35 constructed graphs for each dataset. And then for each constructed graph, we train vanilla GCN \cite{GCN} on it three times and calculate the average test accuracy on node classification task. From the Fig.~\ref{fig:gr_valid}, we find that both the homophily graph and the heterophily graph follow the same rule: when the degree is fixed, the accuracy of GCN increases with the increase of the node-level homophily ratio; when the homophily ratio is fixed, the accuracy of GCN increases with the increase of the degree. It should be noted that when the homophily ratio $p$ equals 0 (i.e., all neighboring nodes are from different classes), it may have a higher GCN accuracy than that when the homophily ratio is very small (around $p=0.2$).  Besides, when the homophily ratio $p$ is largr than a threshold, the GCN accuracy converges to 100\%. In general, the GCN accuracy almost varies monotonically with the node-level homophily ratio and node degree. And this motivates us to use graph rewiring as a way of increasing both node-level homophily ratio and degree.
	
	\begin{figure}[h]
		\begin{minipage}[t]{0.33\linewidth}
			\centering
			\subfloat[Cora.]{\includegraphics[width=\linewidth, height=106pt]{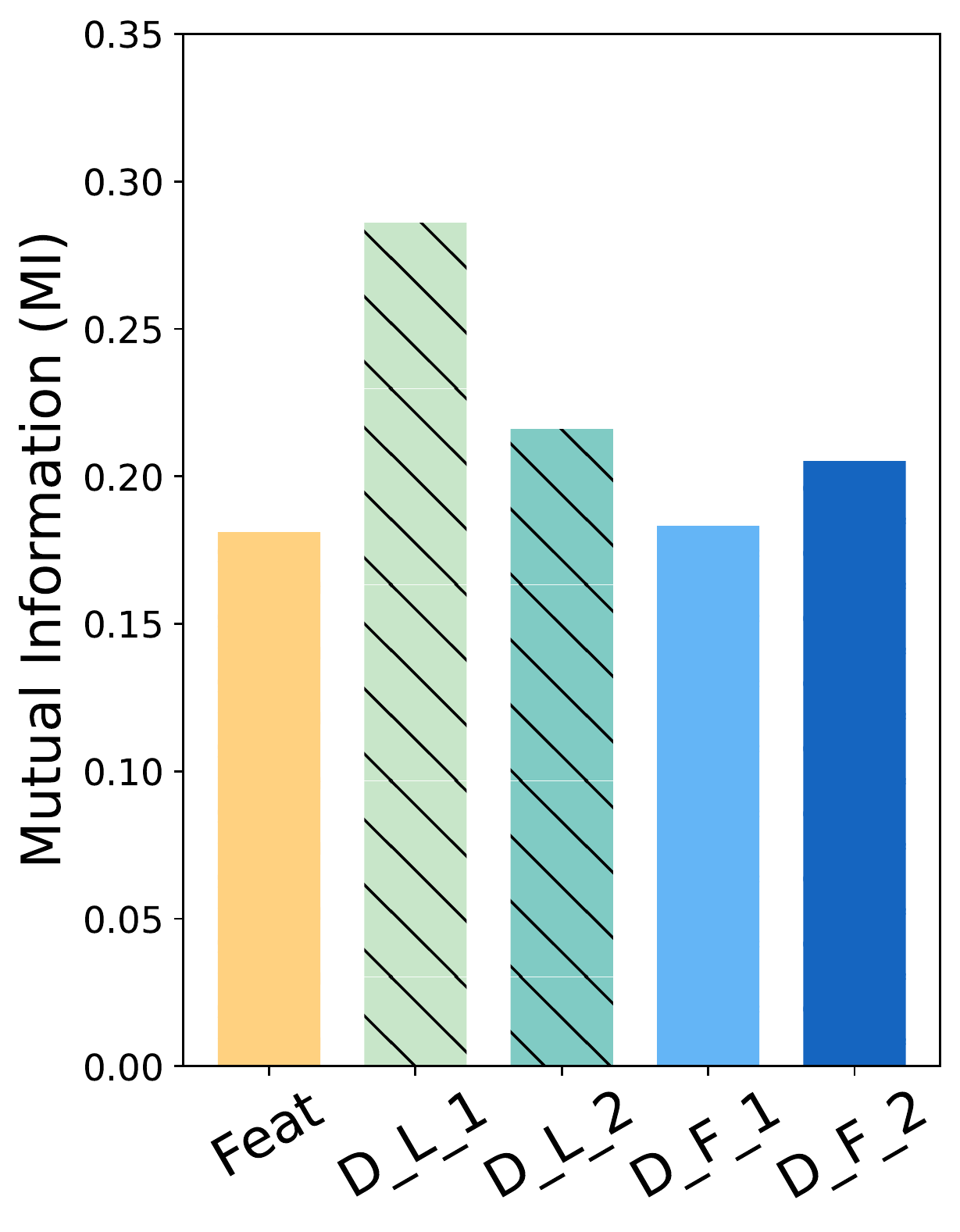}}
		\end{minipage}%
		\begin{minipage}[t]{0.33\linewidth}
			\centering
			\subfloat[Chameleon]{\includegraphics[width=\linewidth, height=106pt]{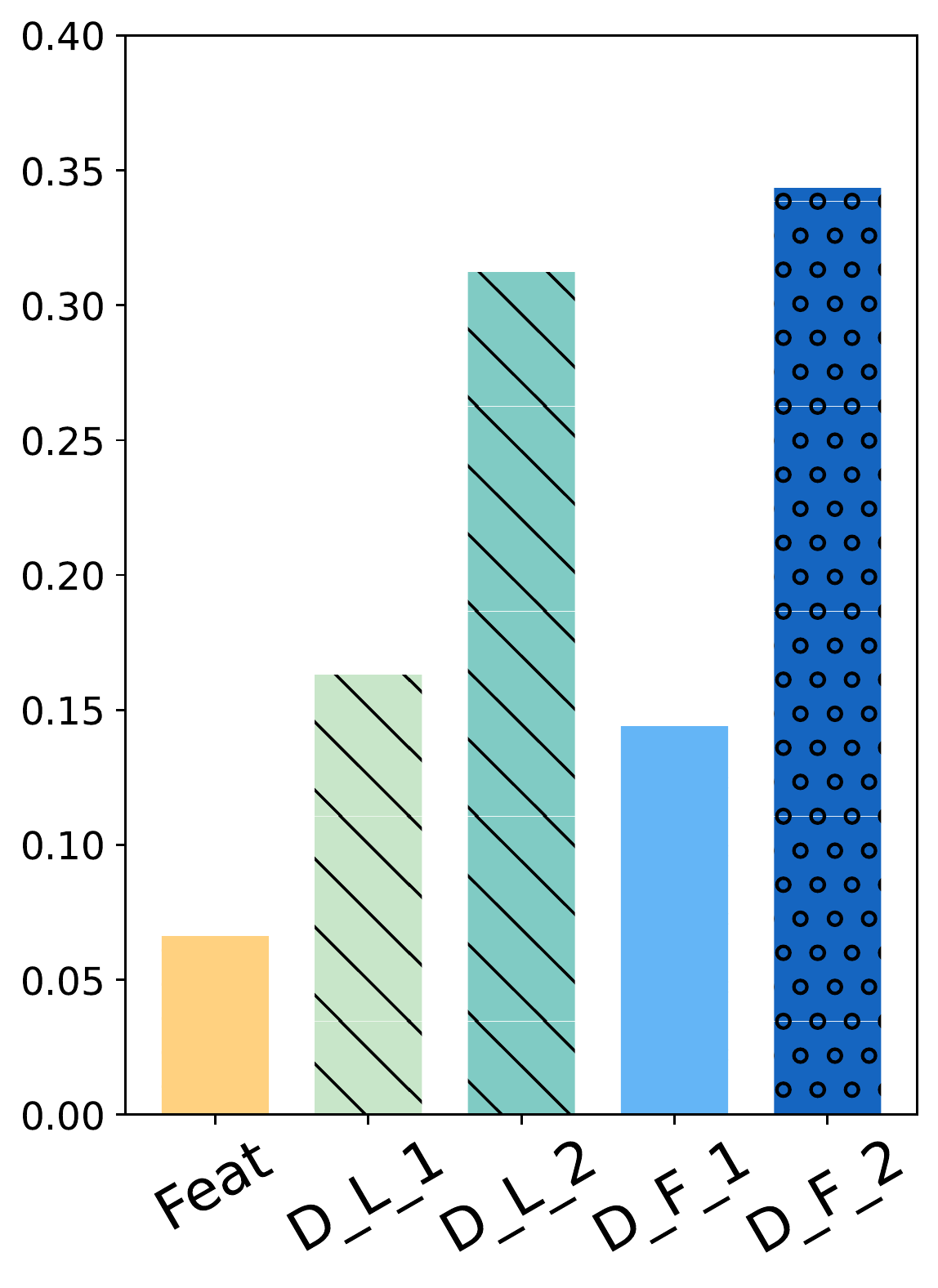}}
		\end{minipage}
		\begin{minipage}[t]{0.33\linewidth}
			\centering
			\subfloat[Actor]{\includegraphics[width=\linewidth, height=106pt]{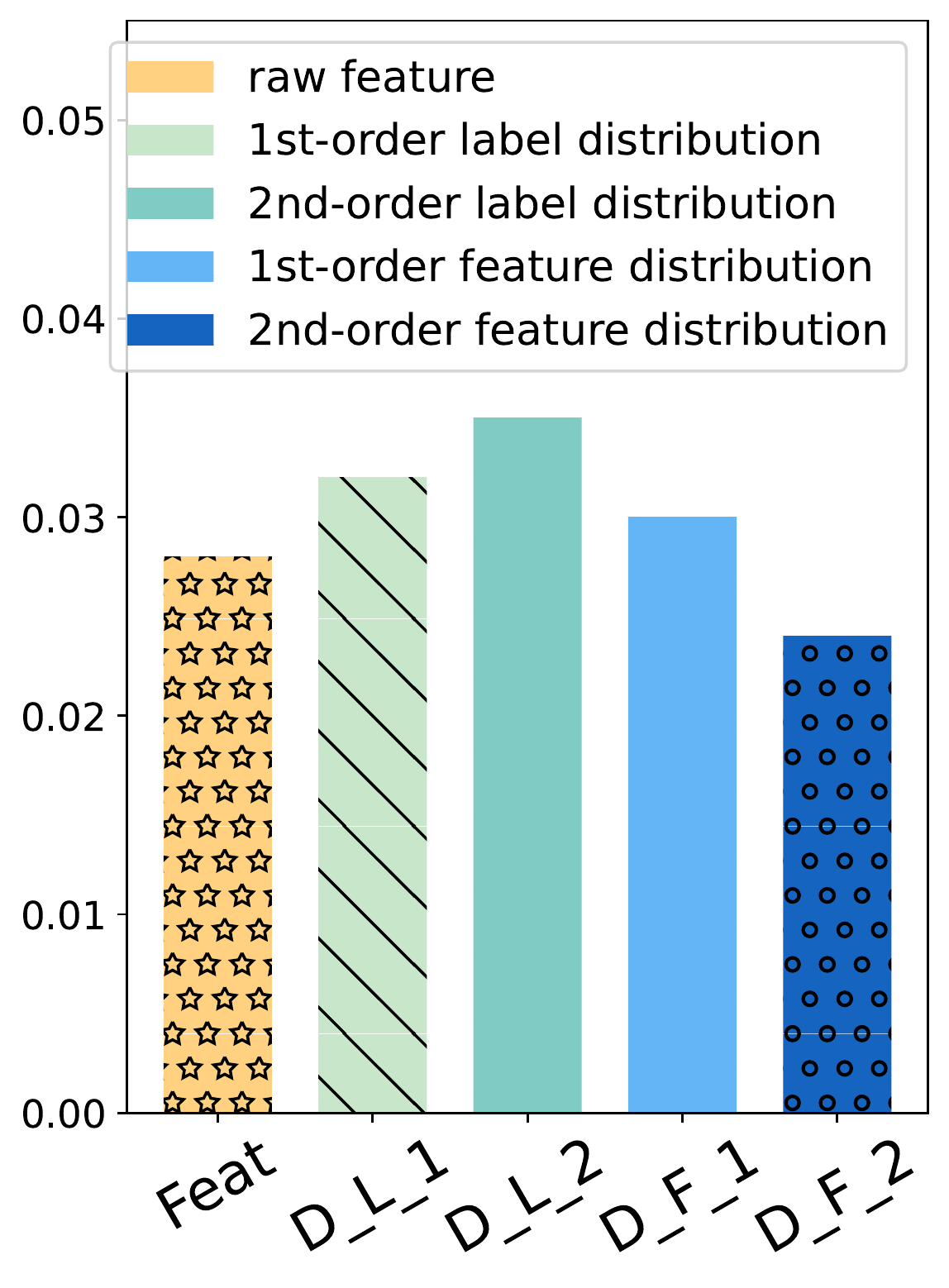}}
		\end{minipage}
		\caption{Mutual Information (MI) between different signals and edge polarity (i.e.homophily or heterophily).}
		
		\label{fig:mi_valid}
	\end{figure}
	\begin{figure*}[h]
		\centering
		\includegraphics[width=0.97\linewidth]{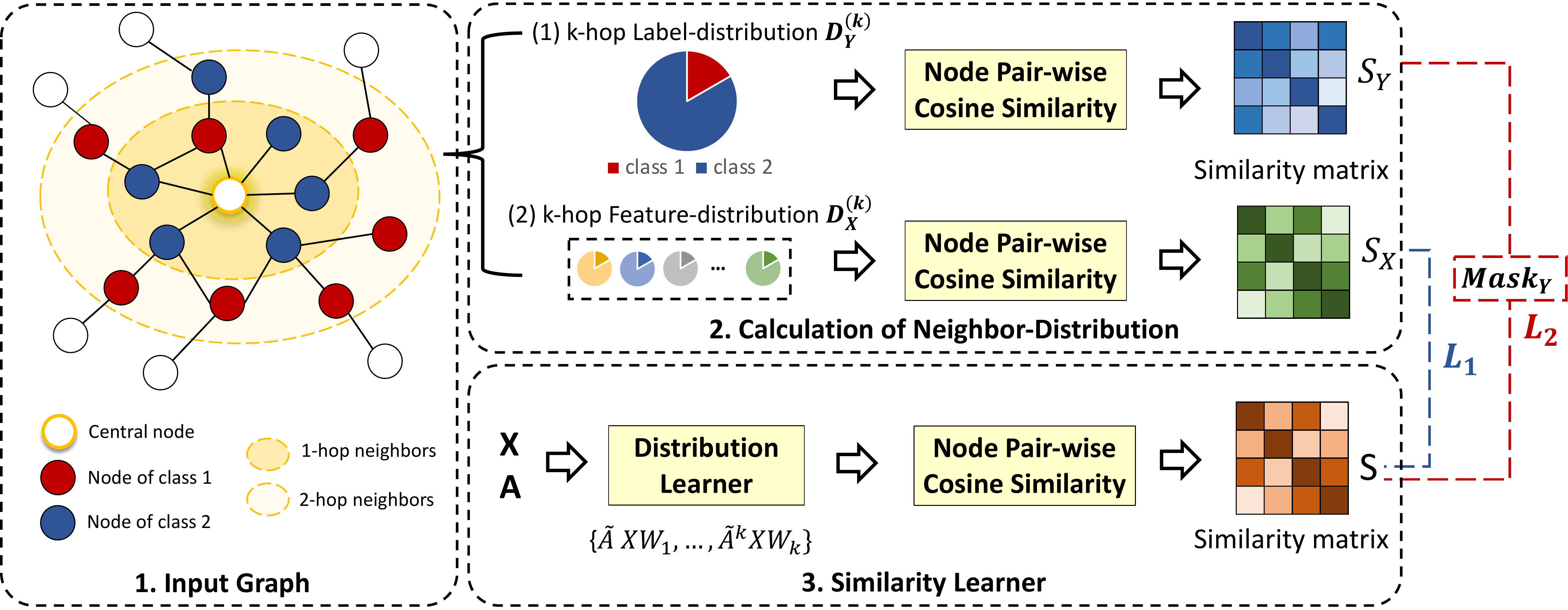}
		\caption{Overview of the Similarity Learner for Graph Rewiring in DHGR. $X\in\mathbb{R}^{N\times D}$ denotes the raw feature matrix and $A\in\mathbb{R}^{N\times N}$ denotes the adjacent matrix. Note that Node Pair-wise Cosine Similarity in the yellow block indicates the cosine similarity with decentralization calculated for each pair of nodes in the graph, which is defined in Eq.~\ref{eq:cos}.}
		\label{fig:model}
	\end{figure*}
	
    \subsection{Effects of Neighbor's Label/Feature Distribution}
	\label{subsec:finding_mi}
	From the Sec.~\ref{subsec:finding_hr_deg}, we conclude that graph rewiring can be used as a way of reducing heterophily to make GNNs perform well on both homophily and heterophily graphs. However, it is not easy to accurately identify the edge polarity (homophily or heterophily) on a heterophily graph so that we can estimate the node-level homophily ratio. For a homophily graph, we can leverage its homophily property and use Laplacian Smoothing \cite{smooth1, smooth2} to make its node representation more distinguishable. However, heterophily graphs do not satisfy property of smoothness thus the information available is limited. A straightforward idea is to use node features to identify edge polarity, however, the information of this single signal is limited.
	In this paper, we propose to use similarity between the \textbf{neighbor's label-distribution} for node-pairs as a measure of edge polarity. Besides, considering that not all node labels are observable, we also introduce \textbf{neighbor's feature-distribution} (mean of neighbor features), which is completely observable, as signals in addition to neighbor's label-distribution. 
	
	Up to now, we have three signals (i.e. raw node features, label-distribution and feature-distribution of neighbors) that can be used as measures for edge polarity. We quantitatively evaluate the effectiveness of the three signals and find that the distribution signals are more informative than the raw node feature through the following empirical experiments and analysis. To be specific, we consider the label/feature-distribution of the 1st-order and 2nd-order neighbors. Then we calculate the similarity between each node-pair with one of these signals and compute the mutual information between the node-pair similarity and edge polarity on the graph.
	The formula of Mutual information is written as follows:
	\begin{equation}
	I(X;Y) = \int_y\int_x P_{(X,Y)}(x,y)\cdot log\left( \frac{P_{X,Y}(x,y)}{P_X(x)P_Y(y)}\ dxdy \right)
	\end{equation}
	In the case of discrete random variables, the integral operation is replaced by the sum operation. 
	
	As shown in the Fig.~\ref{fig:mi_valid}, we conduct statistical analysis on three datasets (i.e. Cora, Chameleon, Actor). From the Fig.~\ref{fig:mi_valid}, we find that both the similarity of neighbor's label-distribution and neighbor's feature-distribution have a strong correlation with edge polarity than that of the raw node features similarity, and neighbor's label-distribution has a stronger correlation than neighbor's feature-distribution in most cases. And this rule applies to both homophily graphs and heterophily graphs.
	\section{Method}
	Based on the observations mentioned above, we design the \textbf{D}eep \textbf{H}eterophily \textbf{G}raph \textbf{R}ewiring method (DHGR) for heterophily graphs, which can be easily plugged into various existing GNN models. Following the pipeline in Fig.~\ref{fig:piepline}, DHGR first learns a similarity matrix representing the similarity between each node-pair based on the neighbor distribution (i.e. label-distribution and feature-distribution of neighbors). Then we can rewire the graph structure by adding edges between high-similarity node-pair and pruning low-similarity edges on the original graph. Finally, the rewired graph is further fed into other GNN models for node classification tasks.

	\subsection{Similarity Learner Based on Neighbor Distribution}
	Before rewiring the graph, we first learn the similarity between each pair of nodes. According to the analysis in Sec.~\ref{subsec:finding_mi}, we design a graph learner that learns the node-pair similarity based on the neighbor distribution. Considering that in the training process, only the labels of nodes in the training set are available, we cannot observe the full label-distribution of neighbors. Therefore, we also leverage the feature-distribution of neighbors which can be fully observed to enhance this similarity learning process with the intuition that node features have correlations to labels for an attributed-graph. Besides, the results shown in Sec.~\ref{subsec:finding_mi} also validate the effectiveness of neighbor's feature-distribution.
	
	The overview of similarity learner used in DHGR is shown in Fig.~\ref{fig:model}. Specifically, for an attributed graph, we can first calculate its observable label-distribution $D_Y^{(k)}$ and feature-distribution $D_X^{(k)}$ for the $k$-hop neighbors of each node using node-labels in the training set and all node-features:
	\begin{equation}
	\label{eq:dist}
	D_Y^{(k)} = (D^{-1}A)^{k}Y^{Train},\quad D_X^{(k)} = (D^{-1}A)^kX
	\end{equation}
	$D_Y^{(k)}$ and $D_X^{(k)}$ is respectively the label-distribution and feature-distribution of $k$-order neighbors in the graph, $M$ is the maximum neighbor-order and we use $M \in \{1,2\}$ in this paper. $Y^{Train}$ is the one-hot label matrix, the $i$-th row of $Y^{Train}$ is the one-hot label vector of node $v_i$ if $v_i$ belongs to the training set, else use a zero-vector instead. $A$ is the adjacent matrix and $D$ is the corresponding degree diagonal matrix. Then for each node, we can get the observed label-distribution vector and feature-distribution vector of its neighbors. Next we calculate the cosine similarity between each node-pair with respect to both label-distribution and feature-distribution, and we can get the similarity matrix of label-distribution $S_Y^{train}$ and similarity matrix of feature-distribution $S_X$.
	\begin{equation}
	\label{eq:x_y_sim}
	\left\{ \begin{aligned}
	&S_Y^{train}[i,j] = \prod_{k=1}^{M} Cos\left({D}_Y^{(k)}[i, :], {D}_Y^{(k)}[j, :]\right) \\
	&S_X[i, j] = \prod_{k=1}^{M} Cos\left({D}_X^{(k)}[i, :], {D}_X^{(k)}[j, :]\right)  \\
	\end{aligned} \right.
	\end{equation}
	where
	\begin{equation}
	\label{eq:cos}
	Cos(z_i, z_j) = \frac{\widetilde{z}_i\cdot \widetilde{z}_j}{|\widetilde{z}_i|\cdot |\widetilde{z}_j|}, \quad \widetilde{z}_i = z_i - \frac{1}{|V|}\sum_{v_j\in V} z_j
	\end{equation}
	Note that before calculating cosine similarity, we first decentralize the input variable by subtracting the mean of this variable for all nodes. Considering that not all nodes have an observed label-distribution, e.g., if all neighbors of node $v_i$ do not belong to the training set, then the observed label-distribution of $v_i$ is a zero vector. Obviously, this is not ideal, so we compensate for this with the feature-distribution of neighbors. In addition, we restrict the utilization condition of neighbor label-distribution by using a \textbf{mask}. Specifically, for node $v_i\in V$, we  leverage its neighbor label-distribution only when the percentage of its neighbors in the training set is larger than a threshold $\alpha$: 
	\begin{equation}
	\label{eq:mask}
	Mask_Y(v_i) = \left\{ 
	\begin{aligned}
	1 &\ , &  r_i > \alpha\\
	0 &\ , & r_i  \leq \alpha\\
	\end{aligned}
	\right.
	,\ \text{where}\ r_i=\frac{|\mathcal{N}(v_i)\cap V^{Train}|}{|\mathcal{N}(v_i)|}
	\end{equation}
	$Mask_Y\in\mathbb{R}^{N\times 1}$ is the mask vector, $\mathcal{N}(v_i)$ is the neighbor set of $v_i$, $V^{Train}$ is the set of nodes in the training set.
	
	Then our similarity learner targets at learning the  similarity of node-pairs based on the neighbor distribution. Specifically, the similarity learner first aggregates and transforms the feature of neighboring nodes and then uses the aggregated node representation to calculate cosine similarity for each node-pair:
	\begin{equation}
	\label{eq:sim_out}
	S_{i,j} = \prod_{k=1}^M Cos\left({h}_i^{(k)}, {h}_j^{(k)} \right), \quad h_i^{(k)} =  (D^{-1}A)^k\cdot X\cdot W
	\end{equation}
	$S_{i,j}$ denotes the similarity between $v_i$ and $v_j$, and similarity of all node-pairs form a similarity matrix $S$. In practice, we also optionally use the concatenation of distribution feature $h_i^{k}$ and  $x_i\cdot W$ (transformed feature of node itself) for similarity calculation in Eq.~\ref{eq:sim_out}. Finally, we use the  $S_X$ and $S_Y^{train}$ calculated in advance to guide the training of $S$. We have the following two objective functions with respect $S_X$ and $S_Y^{train}$:
	\begin{equation}
	\label{eq:loss_pretrain}
	L_{1}(S, S_X) = ||S-S_X||_F^2
	\end{equation}
	\begin{equation}
	\label{eq:loss_finetune}
	L_{2}(S, S_Y^{train}) = ||(S-S_Y^{train})\odot (Mask_Y\times Mask_Y^T)||_F^2
	\end{equation}
	In practice, we first use $S$ to reconstruct $S_X$ as the \textbf{pretraining} process and then further use $S$ to reconstruct $S_Y^{train}$ under $Mask_Y$ as the \textbf{finetuning} process. 
	\subsection{A Scalable Implementation of DHGR}
	However, directly optimizing the objective function mentioned above has quadratic computational complexity. For node attributes $X\in\mathbb{R}^{N\times D}$, the $O(N^2)$ complexity is unacceptable for large graphs when $N>>D$. So we design a scalable training strategy with stochastic mini-batch. Specifically, we randomly select $k_1\cdot k_2$ node-pairs as a batch and optimize the similarity matrix $S$ by a $(k_1\times k_2)$-sized sliding window in each iteration. We can assign small numbers to $k1, k2\in [1, N]$. We give the pseudocode in Algorithm~\ref{algorithm:training}.
	
	\begin{algorithm}
		\caption{Training DHGR with stochastic mini-batch}
		\label{algorithm:training}
		\LinesNumbered
		\KwIn{graph $\mathcal{G}(X,A)$, Node set $V$, label $Y$, batch-size $[k_1, k_2]$, min-percentage $\alpha$ , max neighbor-ordinal $M$,  MaxIteration, $Mask_Y$ Epoch1, Epoch2.} 
		\KwOut{Similarity matrix $S$}
		\For{epoch from 1 to Epoch1}{
			\For{i from 1 to MaxIteration}{
				Sample $k_1$ nodes $V_i^1\in V$, $|V_i^1|=k_1$;\\
				Sample $k_2$ nodes $V_i^2\in V$, $|V_i^2|=k_2$;\\
				Calculate the similarity matrix $\widehat{S}, {\widehat{S}_X}\in \mathbb{R}^{k_1\times k_2}$ between $V_i^1$ and $V_i^2$.   (see Eq.~\ref{eq:sim_out} and Eq.~\ref{eq:x_y_sim});\\
				Update $W$ with  $\nabla L_{1}(\widehat{S}, \widehat{S}_X)$ (see Eq.~\ref{eq:loss_pretrain});\\
			}
		}
		Node set  $V_Y \leftarrow \left\{v_i\in V\ |\ Mask_Y[i]=1 \right\}$.\\
		\For{epoch from 1 to Epoch2}{
			Sample $k_1$ nodes $V_i^1\in V_Y$, $|V_i^1|=k_1$;\\
			Sample $k_2$ nodes $V_i^2\in V_Y$, $|V_i^2|=k_2$;\\
			Calculate the similarity matrix $\widehat{S},{ \widehat{S}_Y^{train}}\in \mathbb{R}^{k_1\times k_2}$ between $V_i^1$ and $V_i^2$. (see Eq.~\ref{eq:sim_out} and Eq.~\ref{eq:x_y_sim});\\
			Update $W$ with the gradient of $ ||(\widehat{S}-\widehat{S}_Y^{train})||_F^2$);\\
		}
		Obtain the entry $S_{i,j}$ with Eq.~\ref{eq:sim_out}. \Comment{Final similarity}
	\end{algorithm}

	\subsection{Graph Rewiring with Learned Similarity}

	After we obtain the similarity of each node-pair, we can use the learned similarity $S$ to rewire the graph. Specifically, we add edges between node-pairs with high-similarity and remove edges with low-similarity on the original graph. Three parameters are set to control this process: $K$ indicates the maximum number of edges that can be added for each node; $\epsilon$ constrains that the similarity of node-pairs to add edges mush larger than a threshold $\epsilon$.  Finally another parameter $\gamma$ is set for pruning edges with similarity smaller than $\gamma$. The details of the Graph Rewiring process are given in Algorithm~\ref{algorithm_GR}. Finally, we can feed the rewired graph $\widehat{\mathcal{G}}(X, \widehat{A})$ into any GNN-based models for node classification tasks.
	
	\begin{algorithm}
		\caption{Graph Rewiring with DHGR}
		\label{algorithm_GR}
		\LinesNumbered
		\KwIn{original graph $\mathcal{G}(V, E)$, learned similarity matrix $S$, max number of added edges $K$, growing threshold $\epsilon$, pruning threshold $\gamma$.} 
		\KwOut{Rewired Graph $\widehat{\mathcal{G}}(X, \widehat{A})$}
		\ForEach{node $v_i\in V$}{
			Select $K$ nodes from $V$ which have top-$K$ largest similarity with $v_i$ to form a node set $C_i$;\\
			Calculate candidate node set $C_i^{'} \leftarrow \left\{ v_j\in C_i\ |\ S_{j,i}\geq \epsilon \right\} $;\\
			Adding edges $\left\{(v_j, v_i)\ |\ v_j\in C_i^{'}\right\}$ to $\mathcal{G}$;\\
		}
		\ForEach{$(v_j, v_i)\in E$}{
			\If{$S_{j, i} < \gamma$}{
				Remove edges $(v_j, v_i)$ from $\mathcal{G}$;\\
			}
		}
	\end{algorithm}
	
	\begin{table*}
		\centering
		\setlength{\tabcolsep}{6.8pt}
		\caption{The stastical information of the datasets used to evaluate our model. H.R. indicates the overall homophily ratio~\cite{pei2020geom} of the dataset, which means the percentage of homophilic edges in all edges of the graph.}
		\label{tab:data_info}
		\begin{tabular}{cccccccccccc}
			\toprule
			Dataset & Chameleon & Squirrel & Actor& FB100 & Flickr  & Cornell & Texas & Wisconsin & Cora & CiteSeer & PubMed  \\
			\midrule
			Nodes & 2277 & 5201 & 7600& 41554 & 89250 & 183 & 183 & 251  & 2708 & 3327 & 19717 \\
			Edges & 36101 & 217073 & 30019 & 2724458 & 899756  & 298 & 325 & 511 & 10556 & 9104 & 88648 \\
			Features & 2325 & 2089 & 93 2& 4814 & 500  & 1703 & 1703 & 1703  & 1433 & 3703  & 500\\
			Classes & 5 & 5 & 5 & 2 & 7 & 5 & 5 & 5 & 7 & 6 &  3\\
			H.R. & 23.5\% & 22.4\% & 21.9\%  & 47.0\%& 31.9\% & 30.5\% & 10.8\% & 19.6\% & 81.0\% & 73.6\% & 80.2\%  \\
			\bottomrule
		\end{tabular}
	\end{table*}

	\subsection{Complexity Analysis}
	We analyze the computational complexity of Algorithm~\ref{algorithm:training} and Algorithm~\ref{algorithm_GR}  with respect to the number of nodes $|V|$. For Algorithm~\ref{algorithm:training}, the complexity of random sampling $k_1+k_2$ nodes is $O(k_1+k_2))$. Lets denote the feature dimension as $D$  and denote the one-hot label dimension as $|\mathcal{C}|< D$. Considering that the complexity of calculating cosine similarity between two $D$-dimension vectors is $O(D)$, the complexity of calculating the similarity matrix $\widehat{S}$, $\widehat{S}_X$, $\widehat{S}_Y^{train}\in\mathbb{R}^{k_1\times k_2}$  is $O(D k_1 k_2)$. 
	The complexity of calculating $L_1$ and $L_2$ equals to $O(k_1 k_2)$.  Therefore, the final computational complexity of one epoch of Algorithm~\ref{algorithm:training} is $\bm{O(D k_1 k_2)}$ where $k_1,k_2$ are two constants.
	For Algorithm~\ref{algorithm_GR}, we use Ball-Tree to compute the top-K nearest neighbors, the complexity of one top-K query is  $O(D\cdot|V| log(|V|)))$. Therefore, the time complexity of the first $FOR$-loop which performs the topK algorithm is approximately $O(D\cdot|V|log(|V|) + K\cdot|V|)$. The second $FOR$-loop filters each edge in the original Graph and thus its complexity is $O(|E|))$. Therefore the final complexity of Algorithm~\ref{algorithm_GR} is $\bm{O(D\cdot|V|log(|V|) + K\cdot|V| + |E|)}$.

	\begin{table*}
		\centering
		\setlength{\tabcolsep}{4.0pt}
		\caption{Node classification accuracy (\%) on the test set of heterophily graph datasets. The bold numbers indicate that our method improves the base model. The dash symbols indicate that we were not able to run the experiments due to memory issue. }
		\label{tab:main_res_hete}
		\begin{tabular}{ccccccccccc}
			\toprule
			GNN Model & $\backslash$ & Chameleon & Squirrel & Actor& Flickr & FB100  & Cornell & Texas & Wisconsin  \\
			\midrule
			
			\multirow{2}*{GCN} & vanilla & 37.68$\pm$3.06 & 26.39$\pm$0.88 & 28.90$\pm$0.57& 49.68$\pm$0.45 & 74.34$\pm$0.20 & 55.56$\pm$3.21 & 61.96$\pm$1.27 & 52.35$\pm$7.07  \\
			~&DHGR & \textbf{70.83$\pm$2.03} & \textbf{67.15$\pm$1.43} & \textbf{36.29$\pm$0.12} & \textbf{51.01$\pm$0.25} & \textbf{77.01$\pm$0.14} & \textbf{67.38$\pm$5.33} & \textbf{81.78$\pm$0.89} &\textbf{ 76.47$\pm$3.62 }\\
			\midrule
			\multirow{2}*{GAT} & vanilla &44.34$\pm$1.42 & 29.82$\pm$0.98 & 29.10$\pm$0.57& 49.67$\pm$0.81 & 70.01$\pm$0.66 & 56.22$\pm$6.02 & 60.36$\pm$5.55 & 49.61$\pm$6.20  \\
			~&DHGR & \textbf{72.11$\pm$2.87} & \textbf{62.37$\pm$1.78} &\textbf{ 34.71$\pm$0.48} & \textbf{50.40$\pm$0.09} & \textbf{79.41$\pm$5.13} & \textbf{70.09$\pm$6.77} & \textbf{83.78$\pm$3.37} & \textbf{73.20$\pm$4.89} \\
			\midrule
			\multirow{2}*{GraphSAGE} & vanilla & 49.06$\pm$1.88 & 36.73$\pm$1.21 & 35.07$\pm$0.15& 50.21$\pm$0.31   & 75.99$\pm$0.09 & 80.08$\pm$2.96 & 82.03$\pm$2.77 & 81.36$\pm$3.91 \\
			~&DHGR& \textbf{69.57$\pm$1.28} & \textbf{68.08 $\pm$1.55} & \textbf{37.17$\pm$0.11}& \textbf{50.85$\pm$0.05}  & \textbf{76.56$\pm$0.10} & \textbf{82.88$\pm$5.56} & \textbf{85.68$\pm$2.72} & \textbf{83.16$\pm$1.72} \\
			\midrule
			\multirow{2}*{APPNP} & vanilla & 40.44$\pm$2.02 & 29.20$\pm$1.45 & 30.02$\pm$0.89 & 49.05$\pm$0.10  & 74.22$\pm$0.11 & 56.76$\pm$4.58 & 55.10$\pm$6.23& 54.59$\pm$6.13 \\
			~&DHGR& \textbf{70.35$\pm$2.62} & \textbf{60.31$\pm$1.51} & \textbf{36.93$\pm$0.86}& \textbf{49.36$\pm$0.05} & \textbf{75.46$\pm$0.11}  & \textbf{68.11$\pm$6.59} & \textbf{81.58$\pm$4.36} & \textbf{77.65$\pm$3.06}  \\
			\midrule
			\multirow{2}*{GCNII} & vanilla& 57.37$\pm$2.35 & 39.51$\pm$1.63 & 31.05$\pm$0.14 & 50.34$\pm$0.22  & 77.06$\pm$0.12 & 61.70$\pm$5.91 & 62.43$\pm$7.37 & 52.75$\pm$4.23 \\
			~&DHGR & \textbf{74.57$\pm$2.56} & \textbf{58.38$\pm$1.79} & \textbf{36.03$\pm$0.12}& \textbf{50.73$\pm$0.31} & \textbf{78.38$\pm$0.91}  & \textbf{72.97$\pm$6.73} & \textbf{81.08$\pm$6.02} & \textbf{78.24$\pm$4.99} \\
			\midrule
			\multirow{2}*{GPRGNN} & vanilla& 41.56$\pm$1.66 & 30.03$\pm$1.11 & 35.72$\pm$0.19 & 49.76$\pm$0.10 & 78.58$\pm$0.23 & 72.78$\pm$6.05 & 69.37$\pm$1.27 & 76.08$\pm$5.86  \\
			~&DHGR & \textbf{71.58$\pm$1.59} &\textbf{ 64.82$\pm$2.07} & \textbf{37.43$\pm$0.78}& \textbf{50.56$\pm$0.32}  & \textbf{82.28$\pm$0.56}  & \textbf{76.56$\pm$5.77} & \textbf{83.98$\pm$2.54} & \textbf{79.41$\pm$4.98} \\
			\midrule
			\multirow{2}*{H2GCN} & vanilla& 49.21$\pm$2.57 & 34.58$\pm$1.61 & 35.61$\pm$0.31  & ---  & ---& 79.06$\pm$6.36 & 80.27$\pm$5.41 & 80.20$\pm$4.51 \\
			~&DHGR & \textbf{69.19$\pm$1.913} & \textbf{72.24$\pm$1.52} & \textbf{36.51$\pm$0.67} & ---  & --- & \textbf{82.06$\pm$6.27} & \textbf{84.86$\pm$5.01} & \textbf{85.01$\pm$5.51}\\
			\midrule 
			Avg Gain & $\backslash$ & 25.51 $\uparrow$ &32.44 $\uparrow$&4.23 $\uparrow$ &0.70 $\uparrow$ &3.15 $\uparrow$  &8.27 $\uparrow$ &15.89 $\uparrow$ &15.17 $\uparrow$ \\
			\bottomrule
		\end{tabular}
	\end{table*}

	\section{Experiments}
	In this section, we first give the experimental configurations, including the introduction of datasets, baselines and setups used in this paper. Then we give the results of experiments comparing DHGR with other graph rewiring methods on the node classification task under transductive learning scenarios. Besides, we also conduct extensive hyper-parameter studies and ablation studies to validate the effectiveness of DHGR.
	
	\subsection{Datasets}
	We evaluate the performanes of DHGR and the existing methods on eleven real-world graphs. To demonstrate the effectiveness of  DHGR , we select eight heterophily graph datasets (i.e. Chameleon, Squirrel, Actor, Cornell, Texas, Wisconsin~\cite{pei2020geom}, FB100~\cite{facebook100}, Flickr  \cite{zeng2019graphsaint}) and three homophily graph datasets (i.e. Cora, CiteSeer, PubMed \cite{GCN}). The detailed information of these datasets are presented in the Table~\ref{tab:data_info}. For graph rewiring methods, we use both the original graphs and the rewired graphs as the input of GNN models to validate their performance on the node classification task.
	\subsection{Baselines}
	DHGR can be viewed as a plug-in module for other state-of-the-art GNN models. And we select five GNN models tackling  homophily, including GCN~\cite{GCN}, GAT~\cite{GAT}, GraphSAGE~\cite{GraphSAGE}, APPNP~\cite{APPNP} and GCNII~\cite{GCN2}. To demonstrate the significant improvement on heterophily graphs caused by DHGR, we also choose two GNNs tackling heterophily (i.e. GPRGNN~\cite{chien2020adaptive}, H2GCN~\cite{h2gcn}). Besides, to validate the effectiveness of DHGR as a graph rewiring method, we also compare DHGR with two Graph Structure Learning (GSL) methods (i.e. LDS~\cite{LDS} and IDGL~\cite{chen2020iterative}) and one Graph Rewiring methods (i.e. SDRF~\cite{topping2021understanding}), which are all aimed at optimizing the graph structure. For GPRGNN and H2GCN, we use the implementation from the benchmark \cite{lim2021new}, and we use the official implementation of other GNNs provided by Torch Geometric. For all Graph Rewiring methods except SDRF whose code is not available, we all use their official implementations proposed in the original papers.
	
	\subsection{Experimental Setup}
	For datasets in this paper, we all use their public released data splits. For Chameleon, Squirrel, Actor, Cornell, Texas ,and Wisconsin, ten random generated splits of data are provided by \cite{pei2020geom}, and we therefore train models on each data split with 3 random seeds for model initialization (totally 30 trails for each dataset) and finally we calculate the average and standard deviation of all 30 results. And we use the official splits of other datasets (i.e. Cora~\cite{GCN}, PubMed~\cite{GCN}, CiteSeer~\cite{GCN}, Flickr~\cite{zeng2019graphsaint}, FB100~\cite{lim2021new}) from the corresponding papers. We train our DHGR models with 200 epochs for pretraining and 30 epochs for finetuning in all datasets. And we search the hyper-parameters of DHGR in the same space for all datasets. $M$ (the max order of neighbors) is searched in \{1, 2\}, $K$ (the growing threshold) is searched in \{3, 6, 8, 16\} and $\gamma$ (the pruning threshold) is searched in \{0., 0.3, 0.6\}, where we do not prune edges for homophily datasets which equals to set $\gamma$ to -1.0. The batch size for training DHGR is searched in \{5000, 10000\}. For other GSL methods (i.e. LDS~\cite{LDS}, IDGL~\cite{chen2020iterative}) , we adjust their hyper-parameters according to the configurations used in their papers. For GNNs used in this paper, we adjust the hyper-parameters in the same searching space for fairness. We search the hidden dimensions in \{32, 64\} for all GNNs and set the number of model layers to 2 for GNNs except for GCNII~\cite{GCN2} which is designed with deeper depth and we search the number of layers for GCN2 in \{2, 64\} according to its official implementation. We train 200/300/400 epoch for all models and select the best parameters via the validation set.  The learning rate is searched in \{1e-2, 1e-3, 1e-4\}, the weight decay is searched in \{1e-4, 1e-3, 5e-3\}, and we use Adam optimizer to optimize all the models on Nvidia Tesla V100 GPU. 

	\subsection{Main Results}
	\label{sec:main_res}
\begin{figure}[h]
		\begin{minipage}[t]{0.33\linewidth}
			\centering
			\subfloat[Cornell.]{\includegraphics[width=\linewidth, height=97pt]{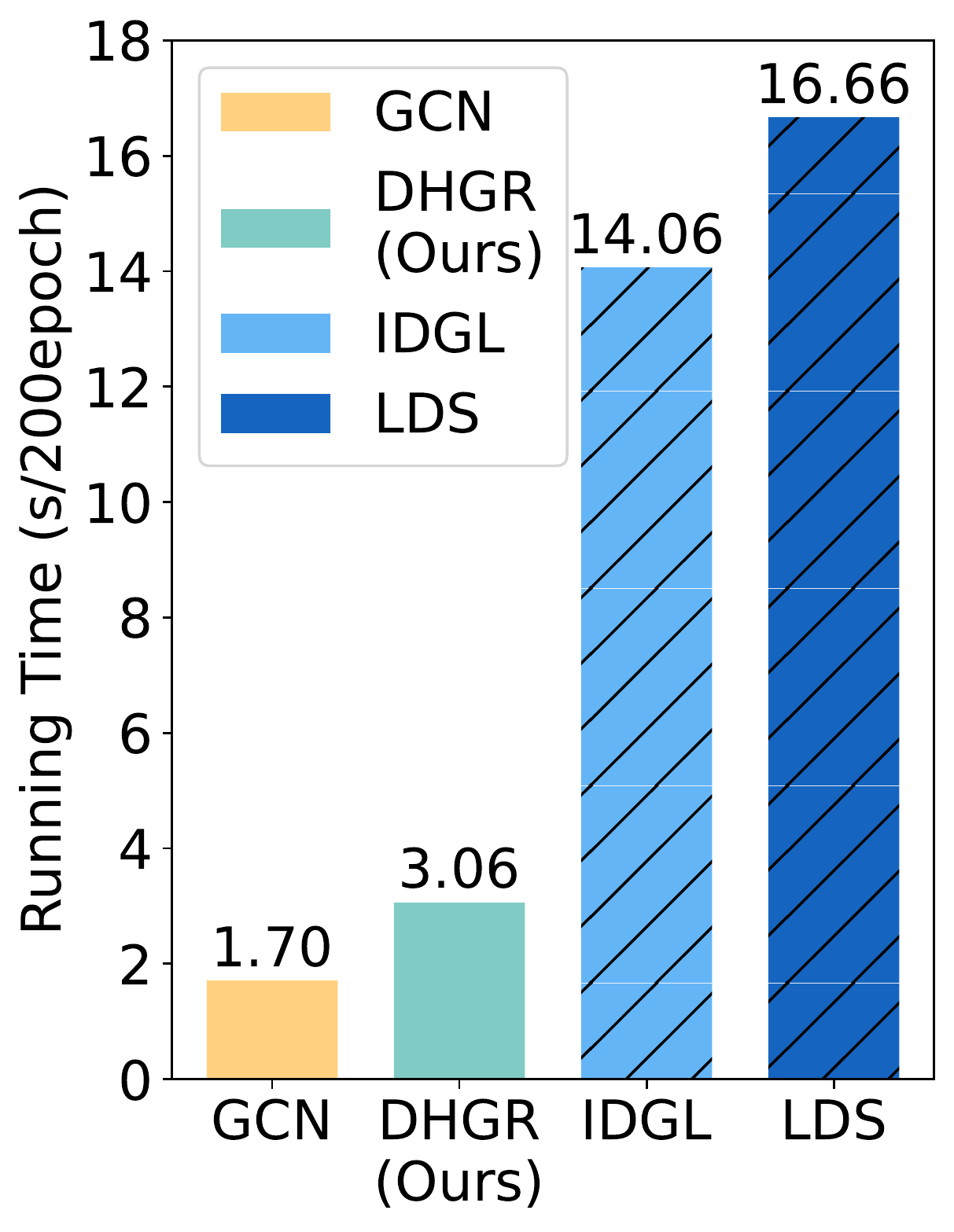}}
		\end{minipage}%
		\begin{minipage}[t]{0.33\linewidth}
			\centering
			\subfloat[Chameleon]{\includegraphics[width=\linewidth, height=97pt]{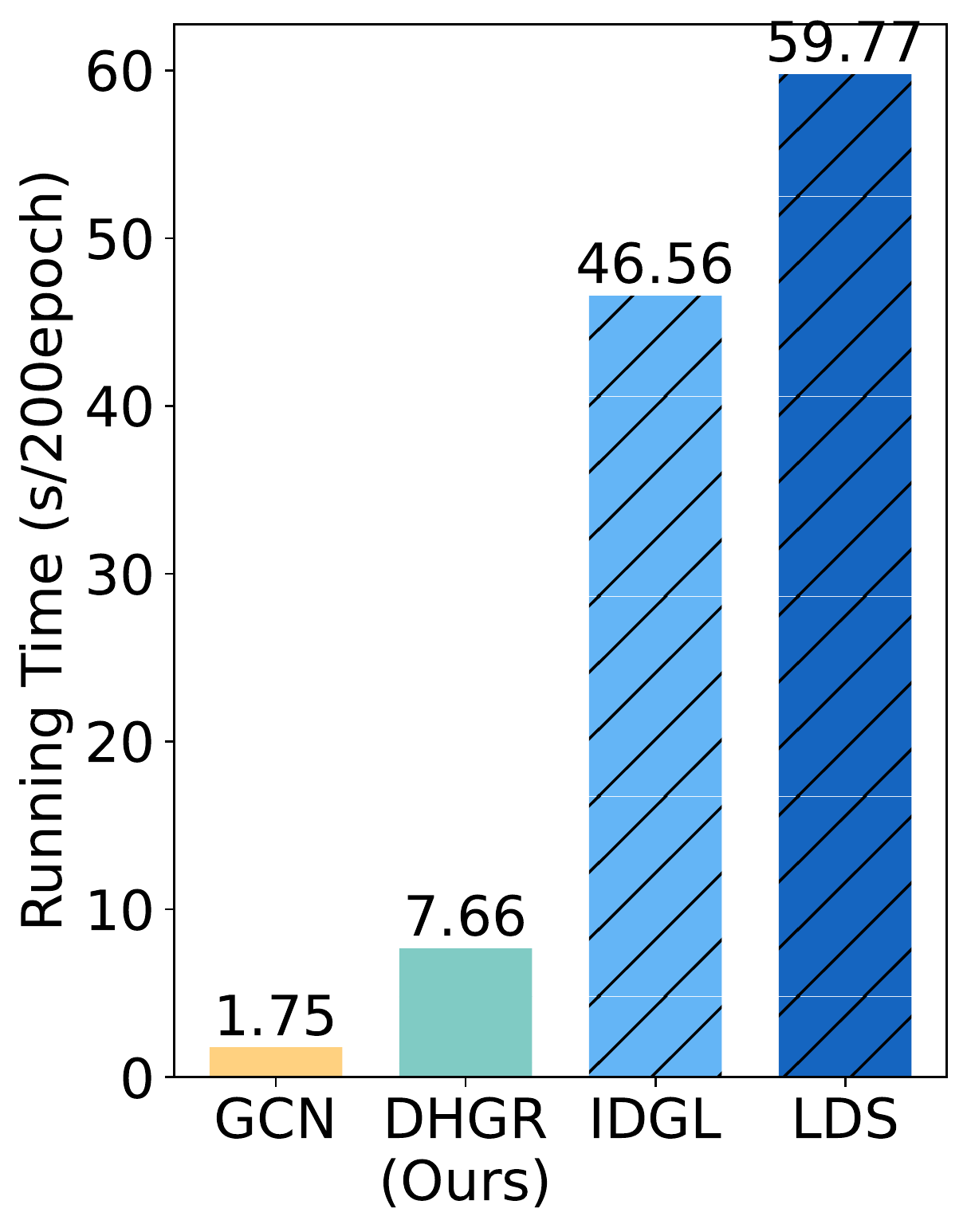}}
		\end{minipage}
		\begin{minipage}[t]{0.33\linewidth}
			\centering
			\subfloat[Actor]{\includegraphics[width=\linewidth, height=97pt]{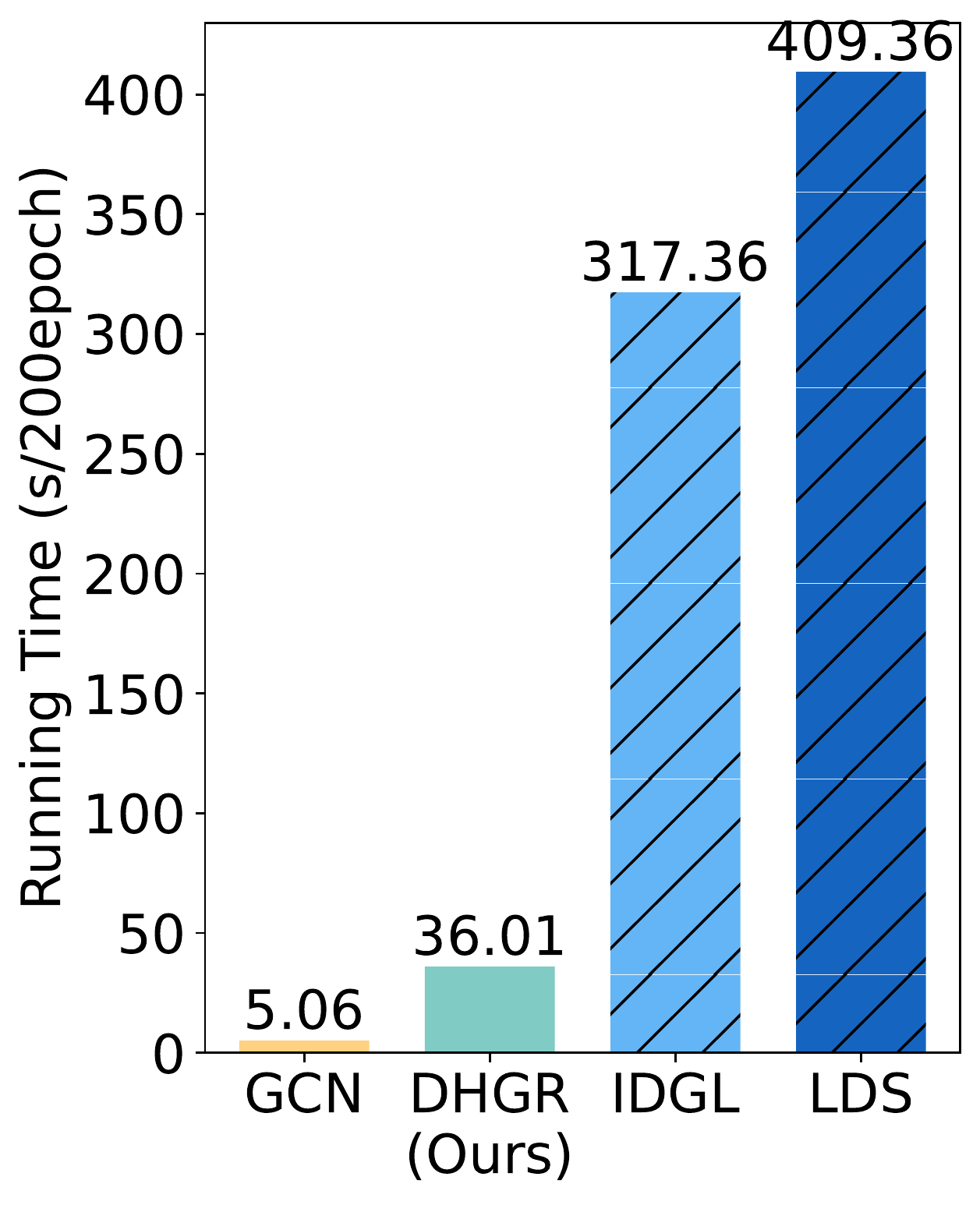}}
		\end{minipage}
		\caption{Running time of GCN with DHGR and other GSL methods (i.e. LDS, IDGL). Note that for DHGR, we use the sum of the running time of DHGR and vanilla GCN as the final running of DHGR for fair comparison with GSL methods. We train 200 epoch for all methods.}
		
		\label{fig:run_time}
	\end{figure}

	We conduct experiments of node classification task on both heterophily and homophily graph datasets, and the results are presented in Table~\ref{tab:main_res_hete} and Table~\ref{tab:main_res_homo} respectively. We evaluate the performance of DHGR by comparing the classification accuracy of GNN with original graphs and graphs rewired by DHGR respectively. We also calculate the average gain (AG) of DHGR for all models on each dataset. The formula of average gain is given as follows:
	\begin{equation}
	AG = \frac{1}{|\mathcal{M}|}\sum_{m_i\in\mathcal{M}} \left( ACC\big(m_i(\widehat{\mathcal{G}})\big) - ACC\big(m_i(\widehat{\mathcal{G}})\big) \right)
	\end{equation}
	where $\mathcal{M}$ is the set of GNN models. $ACC$ is the short form of accuracy.  $\mathcal{G}$ is the original graph and $\widehat{\mathcal{G}}$ is the graph rewired by DHGR.  We also compare the proposed DHGR with other Graph Rewiring methods on their performance and running time, and the results of different graph rewiring methods are reported in Table~\ref{tab:res_g_trans} and Fig.~\ref{fig:run_time}. By analyzing these results,  we have the following observations:
	
	(1) All GNNs enhanced by DHGR, including GNNs for homophily and GNNs for heterophily, outperform their vanilla versions on the eight heterophily graph datasets.  The average gain of DHGR on heterophily graph can be up to 32.44\% on Squirrel. However,  vanilla GCN on Squirrel only has 26.39\% classification accuracy on the test set. Even with the sate-of-the-art GNNs for heterophily (i.e. GPRGNN, H2GCN),  an test accuracy of no more than 40\% can be achieved. The H2GCN enhanced by DHGR can achieve an astonishing 72.24\% test accuracy on Squirrel, almost doubling. For most other heterophily datasets, GNN with DHGR can provide significant accuracy improvements. It demonstrates the importance of graph rewiring strategy for improving GNN's performance on heterophily graphs. Besides, the significant average gain by DHGR also demonstrates the effectiveness of DHGR.  For large-scale and edge-dense datasets such as Flickr and FB100 ($N>>D$), graph rewiring with DHGR can still provide a competitive boost for GNNs, which verifies the effectiveness and scalability of DHGR on large-scale graphs.
	
	\begin{table}
		\centering
		\setlength{\tabcolsep}{5.0pt}
		\caption{Node classification accuracy (\%) on the test set of homophily  graphs. The bold numbers indicate that our method improves the base model.}
		\label{tab:main_res_homo}
		\begin{tabular}{ccccc}
			\toprule
			GNN Model & $\backslash$ & Cora & CiteSeer & PubMed \\
			\midrule
			\multirow{2}*{GCN} & vanilla &81.09$\pm$0.39 & 70.13$\pm$0.45&78.38$\pm$0.39 \\
			~&DHGR & \textbf{82.70$\pm$0.41} & \textbf{70.79$\pm$0.12} & \textbf{79.10$\pm$0.33}  \\
			\midrule
			\multirow{2}*{GAT} & vanilla &81.90$\pm$0.73 &69.60$\pm$0.63 & 78.1$\pm$0.63 \\
			~&DHGR &\textbf{82.93$\pm$0.51} & \textbf{70.43$\pm$0.65} & \textbf{78.81$\pm$0.93} \\
			\midrule
			\multirow{2}*{GraphSAGE} & vanilla & 80.62$\pm$0.47 & 70.30$\pm$0.57 & 77.1$\pm$0.23 \\
			~&DHGR & \textbf{81.30$\pm$0.26} & \textbf{71.11$\pm$0.65} & \textbf{77.63$\pm$0.16}\\
			\midrule
			\multirow{2}*{APPNP} & vanilla & 83.25$\pm$0.42 & 70.46$\pm$0.31 & 78.9$\pm$0.45 \\
			~&DHGR & \textbf{83.86$\pm$0.40} & \textbf{71.60$\pm$0.35} & \textbf{79.61$\pm$0.53}\\
			\midrule
			\multirow{2}*{GCNII} & vanilla & 83.11$\pm$0.37 & 70.90$\pm$0.73 & {79.46$\pm$0.33} \\
			~&DHGR & \textbf{83.93$\pm$0.28} & \textbf{71.96$\pm$0.67} &\textbf{79.49$\pm$0.39}\\
			\midrule 
			Avg Gain & $\backslash$ & 0.95 $\uparrow$ &0.90 $\uparrow$&0.54 $\uparrow$\\
			\bottomrule
		\end{tabular}
	\end{table}
	(2) For homophily graphs (i.e., Cora, Citeseer, Pubmed), the proposed DHGR can still provide competitive gain of node classification performance for the GNNs. Note that homophily graphs usually have a higher homophily ratio (i.e. 81\%, 74\%, 80\% for Cora, CiteSeer and PubMed),  so even vanilla GCNs can achieve great results and thus the benefit of adjusting the graph structure to achieve a higher homophily ratio is less than that for heterophily graphs. To be specific, DHGR gains best average gain on Cora, e.g., the classification accuracy of vanilla GCN on Cora is improved from 81.1\% to 82.6\%. For another two datasets, DHGR also provide average gain no less than 0.5\% accuracy for all GNN models. These results demonstrate that our method can provide significant improvements for heterophily graphs while maintaining competitive improvements for homophily graphs.
	
	\begin{table}
		\centering
		\setlength{\tabcolsep}{1pt}
		\caption{Node classification accuracy (\%) of GCN with different graph rewiring methods. Model with * means we use the results from the original paper (under the same settings of datasets) for their code is unavailable.  The bold numbers indicate that our method improves the base model.}
		\label{tab:res_g_trans}
		\begin{tabular}{ccccc}
			\toprule
			Methods  & Chameleon & Squirrel & Actor & Texas \\
			\midrule
			Vanilla GCN & 37.68$\pm$3.06 & 26.39$\pm$0.88 & 28.90$\pm$0.57 & 61.96$\pm$1.27 \\
			RandAddEdge & 32.17$\pm$6.06 & 22.77$\pm$5.05 & 26.68$\pm$2.26 & 55.85$\pm$1.68 \\
			RandDropEdge & 39.01$\pm$2.47 & 26.48$\pm$1.09 & 29.54$\pm$0.36 &66.76$\pm$1.52 \\
			$\text{RandAddEdge}_\text{Y}^\text{train}$ & 37.01$\pm$3.36 & 27.89$\pm$2.28 & 29.57$\pm$1.17 & 60.08$\pm$2.13 \\
			LDS & 36.12$\pm$2.89 & 28.02$\pm$1.78 & 27.58$\pm$0.97 & 58.75$\pm$5.57 \\
			IDGL & 37.28$\pm$3.36 & 23.57$\pm$2.07 & 27.17$\pm$0.85 & {67.57$\pm$5.85} \\
			SDRF* & \underline{44.46$\pm$0.17} & \underline{41.47$\pm$0.21} & \underline{29.85$\pm$0.07} & \underline{70.35$\pm$0.60}\\
			DHGR & \textbf{70.83$\pm$2.03} & \textbf{67.15$\pm$1.43} & \textbf{36.29$\pm$0.12} & \textbf{81.78$\pm$0.89} \\
			\bottomrule
		\end{tabular}
	\end{table}

	(3) To demonstrate the effectiveness of DHGR as a method of graph rewiring, we also compare the proposed approach with other graph rewiring methods (i.e. LDS, IDGL, SDRF). Besides, we also use two random graph structure transformation by adding or removing edges on the original graph with a probability of 0.5, namely RandAddEdge and RandDropEdge. To validate the effect of adding edges between same-class nodes with training label, we also design a method named $\text{RandAddEdge}_\text{Y}^\text{train}$ that randomly adds edges between same-class nodes \textbf{within the training set} (for we can only observe labels of node in the training set) with a probability of 0.5. As shown in Table~\ref{tab:res_g_trans}, GCN with DHGR outperform GCN with other graph transformation methods on the presented four heterophily datasets. Note that $\text{RandAddEdge}_\text{Y}^\text{train}$ which only use training label to add edges, though increases the homophily ratio, it cannot add edges beyond nodes in the training set. Only adding homophilic edges within the training set cannot guarantee an improvement of GCN's performance and make the nodes in the training set easier to distinguish, increasing the risk of overfitting. The significant improvements made by DHGR demonstrates the effectiveness of DHGR  as a  graph rewiring method. 
	
	(4) Note that the traditional paradigm of GSL methods (e.g., LDS, IDGL.) is training a graph learner and  a GNN through an end2end manner and based on the dense matrix optimization, which have larger complexity. The running time of DHGR and two other GSL methods is presented in Fig.~\ref{fig:run_time}, we find that the running time of DHGR is significantly smaller than that of GSL methods under the same device environment. We did not present the running time of SDRF because its code has not been released publicly yet.
	
	\subsection{Hyper-Parameter Study}
	To demonstrate the robustness of the proposed approach, we study the effect of the four main hyper-parameter of DHGR, i.e. Batchsize,   $K$ (maximum number of added edges for each node), $\epsilon$ (the threshold of lowest-similarity when adding edges) and training ratio of datasets in this section.
	\begin{table}
		\centering
		\setlength{\tabcolsep}{5.0pt}
		\caption{Node classification of GCN enhanced by DHGR with different training ratio and batch size. For each dataset under certain training ratio, we randomly generate 3 data splits and calculate the average accuracy. 
		}
		\label{tab:res_bs}
		\begin{tabular}{ccccccc}
			\toprule
			Dataset & \multicolumn{3}{c}{Squirrel} & \multicolumn{3}{c}{FB100}  \\
			\midrule
			Batchsize &$40\%$&$20\%$& $10\%$&$40\%$&$20\%$& $10\%$\\
			\midrule
			100$\times$100 & 64.57 & 64.01 & 63.31 & 75.36 & 75.02 & 74.78  \\
			1000$\times$1000 & 66.01 & 65.68 & 64.53  & 76.21 & 76.30 & 75.01  \\
			5000$\times$5000 & \underline{66.57} & \underline{66.21} & \underline{66.17} & 76.58 & 76.37 & 75.97  \\
			10000$\times$10000 & \textbf{67.79} & \textbf{67.66} & \textbf{66.32} &\textbf{ 77.32} & \underline{76.57} & \textbf{76.32}   \\
			$N\times N$ & \textbf{67.79} & \textbf{67.66} & \textbf{66.32}  & \underline{77.23} & \textbf{76.87} & \underline{76.21 } \\
			\bottomrule
		\end{tabular}
		
	\end{table}
	\subsubsection{The effect of batchsize and training set ratio}
	Table~\ref{tab:res_bs} shows the results of GCN with DHGR on two heterophily datasets varying with different batchsize for DHGR and training ratio (percentage of nodes in the training set.). The batchsize is ranging from $[100\times 100]$ to $[N\times N]$, where $N$ is the number of nodes and $[N\times N]$ indicates using full-batch for training. Note that for the Squirrel dataset which has only 5201 nodes, the batchsize of $10000\times 10000$ equals full-batch. The results in Table~\ref{tab:res_bs} show that the proposed approach has stable improvements with different batchsize and training ratio. To be specific, GCN with DHGR only has a 3\% decrease in accuracy when decreasing the batchsize to 100, which is extremely small and with no more than 2\% decrease in accuracy with training ratio ranging from 40\% to 10\%. Besides, we usually set the batchsize of DHGR ranging from 5000 to 10000 in real applications, because the overhead of 10000$\times$10000 matrix storage and operation is completely acceptable. These results demonstrate the robustness of DHGR when adjusting the batchsize and training ratio.

	\begin{figure}[h]
		\begin{minipage}[t]{0.5\linewidth}
			\centering
			\subfloat[Homophily Ratio of rewired graphs]{\includegraphics[width=\linewidth]{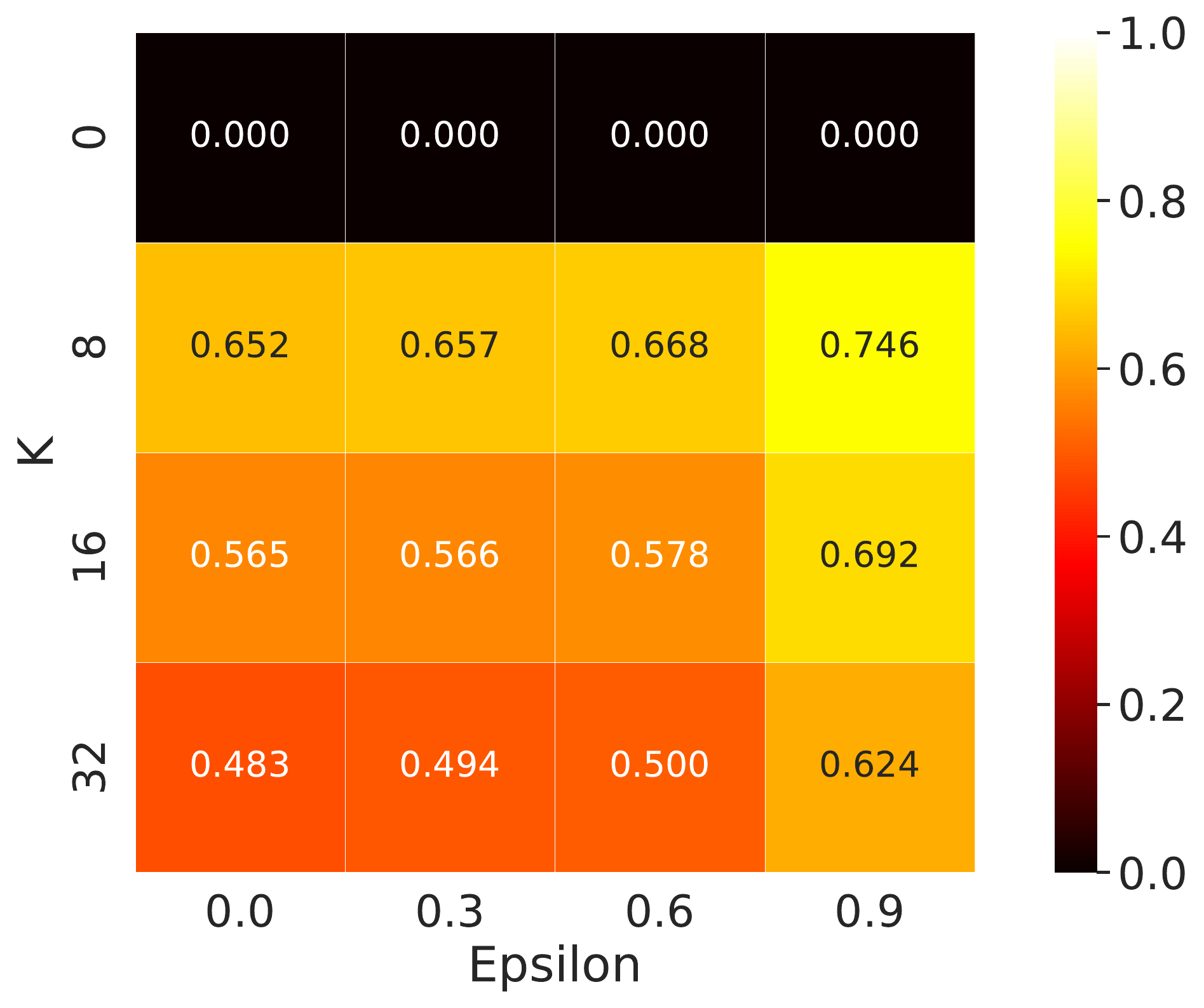}}
		\end{minipage}%
		\begin{minipage}[t]{0.5\linewidth}
			\centering
			\subfloat[GCN Accuracy on rewired graphs]{\includegraphics[width=\linewidth=]{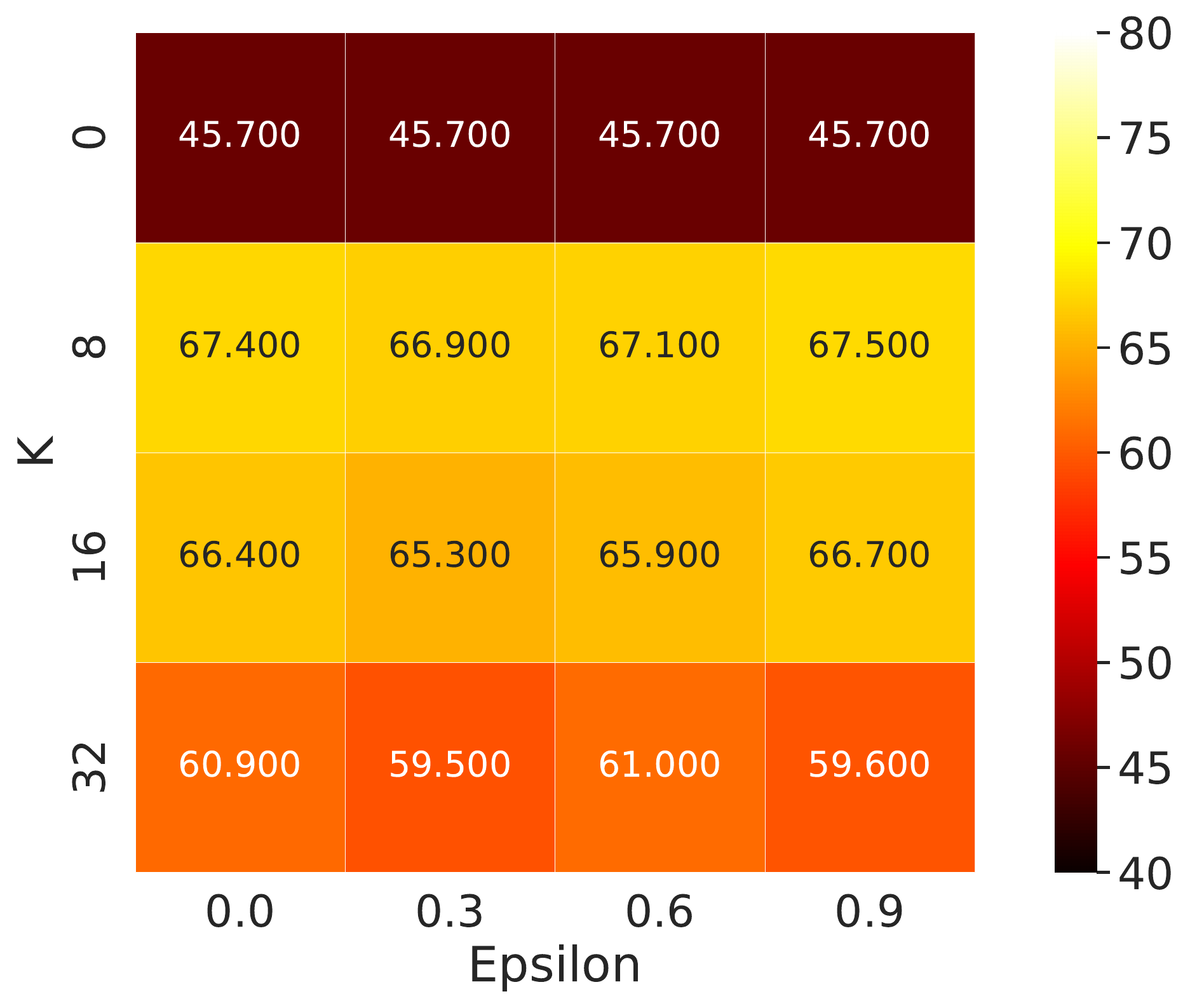}}
		\end{minipage}
		\caption{Results of experiments with different $K$ and $\epsilon$. $K$ is the maximum number of edges that can be added for each node. $\epsilon$ is the minimum similarity threshold of node-pairs between which edges can be added. Note that we remove all edges on the original graph for this experiment and only to verify the effects of edges added by DHGR.}
		
		\label{fig:heat_k_epsilon}
	\end{figure}
	\subsubsection{The effect of  $K$ and $\epsilon$}
	We have two important hyper-parameters when rewiring graphs with DHGR, the maximum number of edges added for each node (denoted as  $K$) and the threshold of lowest-similarity when adding edges (denoted as $\epsilon$). Given the learned similarity by DHGR, the two hyper-parameters almost determines the degree and homophily ratio of the rewired graph. Motivated by the obversations presented in Sec.~\ref{section:findings}, we verify the effectiveness of DHGR for graph rewiring by using different $K$ and $\epsilon$. Fig.~\ref{fig:heat_k_epsilon}~(a) shows the homophily ratio of rewired graphs using different $K$ and $\epsilon$ and Fig.~\ref{fig:heat_k_epsilon}~(b) shows the node classification accuracy of GCN on the rewired graphs using different $K$ and $\epsilon$. We observe that the homophily ratio usually increases when increasing $\epsilon$ with fixed $K$, while decreases when increasing $K$ with fixed $\epsilon$. Besides, the change of GCN node classification accuracy basically matches the change of homophily ratio with different $K$ and $\epsilon$. This demonstrates the effectiveness and robustness of the rewired graphs learned by DHGR.
	\subsection{Ablation Study}
	Considering that DHGR leverage three different types of information (i.e. raw feature, label-distribution, feature-distribution), we also verify the effectiveness of each type of formation by removing them from DHGR and designing three variants of it. $\text{DHGR}_{\backslash \text{label\_dist}}$ indicates removing the using of neighbor's label-distribution (the finetuning process). $\text{DHGR}_{\backslash \text{feat\_dist}}$ indicates removing the using of neighbor's feature-distribution (the pretraining process).
	$\text{DHGR}_{\backslash \text{feat\_self}}$ indicates do not use the concatenation of distribution feature $h_i^{k}$ and  $x_i\cdot W$ (transformed feature of node itself) for similarity calculation in Eq.~\ref{eq:sim_out} (only use the distribution feature $h_i^{k}$). As shown in Table~\ref{tab:ablation}, the node classification of GCN with rewired graphs from almost all variants deteriorates to some extent on the four selected datasets (i.e. Cora, Cornell, Texas, FB100). For the Texas dataset, the results of $\text{DHGR}_{\backslash \text{feat\_dist}}$  that do not utilize neighbors feature-distribution have slight improvement over the full DHGR and we think it is caused by the poor performance of feature-distribution reflected by the results of $\text{DHGR}_{\backslash \text{label\_dist}}$, which only leverages the feature-distribution and feature of node itself  on this dataset. And the result of DHGR on Texas dataset only decreases slightly with 0.2\% accuracy compared with $\text{DHGR}_{\backslash \text{feat\_dist}}$. The results of the ablation study demonstrate the effectiveness of neighbor label-distribution for modeling heterophily graphs. Also, it demonstrates that the proposed approach makes full use of the useful information from  neighbor distribution and raw feature.
	\begin{table}
		\small
		\centering
		\setlength{\tabcolsep}{3.0pt}
		\caption{Node classification accuracy (\%) of the ablation studies to compare GCN with DHGR and its variants which remove certain component from the original DHGR architecture.
		}
		\label{tab:ablation}
		\begin{tabular}{ccccc}
			\toprule
			Methods  & Cora & Cornell & Texas & FB100 \\
			\midrule
			$\text{DHGR}_{\backslash \text{label\_dist}}$ & 80.97$\pm$0.05 & {65.38$\pm$5.53} & {79.67$\pm$1.79}  & {75.95$\pm$0.16}  \\
			$\text{DHGR}_{\backslash \text{feat\_dist}}$ & 81.3$\pm$0.13  & \underline{67.08$\pm$6.08} & \textbf{82.02$\pm$1.06}  & {76.68$\pm$0.56}  \\
			$\text{DHGR}_{\backslash \text{feat\_self}}$ & \underline{81.7$\pm$0.11 }& 62.21$\pm$4.49 & 67.85$\pm$1.02 & \underline{75.65$\pm$0.26}  \\
			DHGR & \textbf{82.63$\pm$0.41}  & \textbf{67.38$\pm$5.33} & \underline{81.78$\pm$0.89}   & \textbf{77.01$\pm$0.14}  \\
			\bottomrule
		\end{tabular}
	\end{table}
	\section{Related work}
	\subsection{Graph Representation Learning}
	Graph Neural Networks (GNNs) have been popular for modeling graph data \cite{bi2022mm,yang2021domain,chen2021fast,wang2019tag2gauss,du2022understanding}. GCN~\cite{GCN} proposed to use graph convolution based on neighborhood aggregation.  GAT~\cite{GAT} proposed to use attention mechanism to learn weights for neighbors. GraphSAGE~\cite{GraphSAGE} was proposed with graph sampling for inductive learning on graphs. These early methods are designed for homophily graphs, and they perform poorly on heterophily graphs. Recently, some studies \cite{abu2019mixhop, pei2020geom, h2gcn, chien2020adaptive, du2022gbk} propose to design GNNs for modeling heterophily graphs. MixHop~\cite{abu2019mixhop} was proposed to aggregate representations from multi-hops neighbors to alleviate heterophily. Geom-GCN~\cite{pei2020geom}  proposed a bi-level aggregation scheme considering both node embedding and structural neighborhood.  GPR-GNN\cite{chien2020adaptive} proposed to adaptively learn the Generalized PageRank (GPR) weights to jointly optimize node feature and structural information extraction. More recently, GBK-GNN~\cite{du2022gbk} was designed with bi-kernels for homophilic and heterophilic neighbors respectively.
	\subsection{Graph Rewiring}
	The traditional message passing GNNs usually assumes that messages are propagated on the original graph \cite{GCN,GAT,GraphSAGE,GCN2}. Recently, there is a trend to decouple the input graph from the graph used for message passing.  For example,  graph sampling methods for inductive learning \cite{GraphSAGE, zhang2019bayesian},  motif-based methods \cite{monti2018motifnet} or graph filter leveraging multi-hop neighbors \cite{abu2019mixhop}, or changing the graph either as a preprocessing step \cite{klicpera2019diffusion,alon2020bottleneck} or adaptively for the downstream task \cite{kazi2022differentiable,wang2019dynamic}. Besides, Graph Structure Learning (GSL) methods \cite{li2018agcn, LDS, chen2020iterative, zhusurvey, gao2020exploring, wan2021graph} aim at learning an optimized graph structure and its corresponding node representations jointly. Such methods of changing graphs for better performance of downstream tasks are often generically named as \textbf{graph rewiring} \cite{topping2021understanding}.
	The works of \cite{alon2020bottleneck,topping2021understanding} proposed rewiring the graph as a way of reducing the bottleneck, which is a structural property in the graph leading to over-squashing.  Some GSL methods \cite{wan2021graph, gao2020exploring}  directly make adjacent matrix a learnable parameter and optimize it with GNN. Other GSL methods \cite{LDS, chen2020iterative} use a bilevel optimization pipeline, in which the inner loop denotes the downstream tasks and the outer loop learns the optimal graph structure with a structure learner. Some studies~\cite{ying2021transformers, dwivedi2021graph} also use transformer-like GNNs to construct global connections between all nodes. However, both GSL methods and graph  transformer-based methods usually have a higher time and space complexity than other graph rewiring methods. Most of existing Graph Rewiring methods are under the similar assumption (e.g., sparsity~\cite{sparse}, low-rank~\cite{lowrank}, smoothness~\cite{smooth1, smooth2}) on graphs. However, the property of low-rank and smoothness are not satisfied by heterophily graphs.  Thus, graph rewiring methods for modeling heterophily graphs still need to be explored.
\section{Conclusion}
In this paper, we propose a new perspective of modeling heterophily graphs by graph rewiring, which targets at improving the homophily ratio and degree of the original graphs and making GNNs gain better performance on the node classification task. Besides, we design a learnable plug-in module  of graph rewiring  for heterophily graphs namely DHGR which can be easily plugged into any GNN models to improve their performance on heterophily graphs. DHGR improves homophily of graph by adjusting structure of the original graph based on neighbor's label-distribution.  And we design a scalable optimization strategy for training DHGR  to guarantee a linear computational complexity. Experiments on eleven real-world datasets demonstrate that DHGR can provide significant performance gain for GNNs under heterophily, while gain competitive performance under homophily. The extensive ablation studies further demonstrate the effectiveness of the proposed approach.
\bibliographystyle{ACM-Reference-Format}
\balance
\bibliography{main}
\appendix
\end{document}